%% file: main.tex
\documentclass{article}
\usepackage{iclr2026_conference,times}

\input{math_commands.tex}

\usepackage[table,dvipsnames]{xcolor}

\usepackage{graphicx}
\usepackage{xcolor}         % 现在会带上上面传的选项
\usepackage{booktabs}
\usepackage{multirow}       % 只保留一次
\usepackage{enumitem}
\usepackage{adjustbox}
\usepackage{pifont}
\usepackage{wrapfig}
\usepackage{array}
\usepackage{float}
\usepackage{xspace}

\usepackage{hyperref}
\usepackage{cleveref}

\crefname{figure}{Fig.}{Figs.}
\Crefname{figure}{Fig.}{Figs.}
\crefname{table}{Tab.}{Tabs.}
\Crefname{table}{Tab.}{Tabs.}

\newcommand{\modelname}{\ensuremath{\mathcal{F}_1}\xspace}
\definecolor{myred}{rgb}{0.945,0.635,0.608}

\title{\includegraphics[height=8mm]{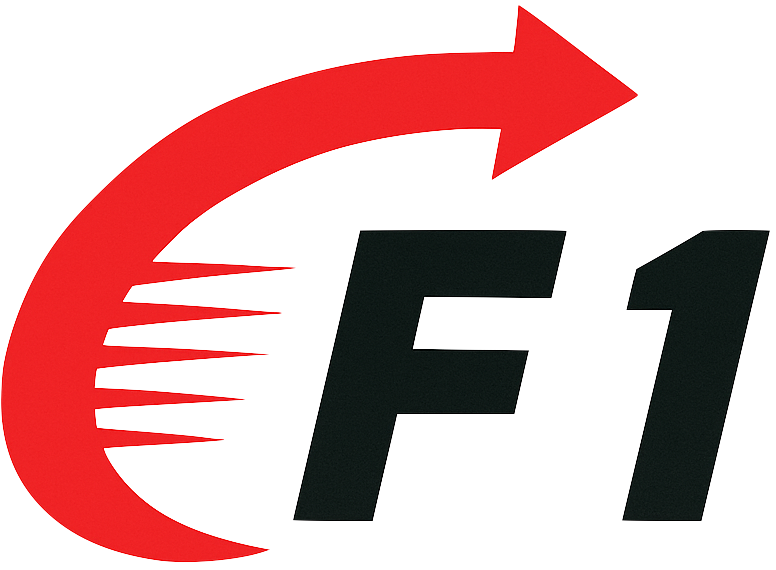}: A Vision-Language-Action Model Bridging Understanding and Generation to Actions}

\author{Qi Lv$^{1,2,*}$, Weijie Kong$^{1,*}$, Hao Li$^{1,2,*}$, Jia Zeng$^{1,\dagger}$, Zherui Qiu$^{1}$, Delin Qu$^{1}$, Haoming Song$^{1}$, \\
\textbf{Qizhi Chen$^{1}$, Xiang Deng$^{2}$, Jiangmiao Pang$^{1,\dagger}$} \vspace{1mm}\\
$^{1}$Shanghai AI Laboratory\quad $^{2}$Harbin Institute of Technology (Shenzhen)\\
$^*$Equal Contributions, $^\dagger$Corresponding Authors\\
\texttt{\{zengjia, pangjiangmiao\}@pjlab.org.cn} \vspace{2mm}\\
\textbf{\includegraphics[height=1.2\fontcharht\font`X]{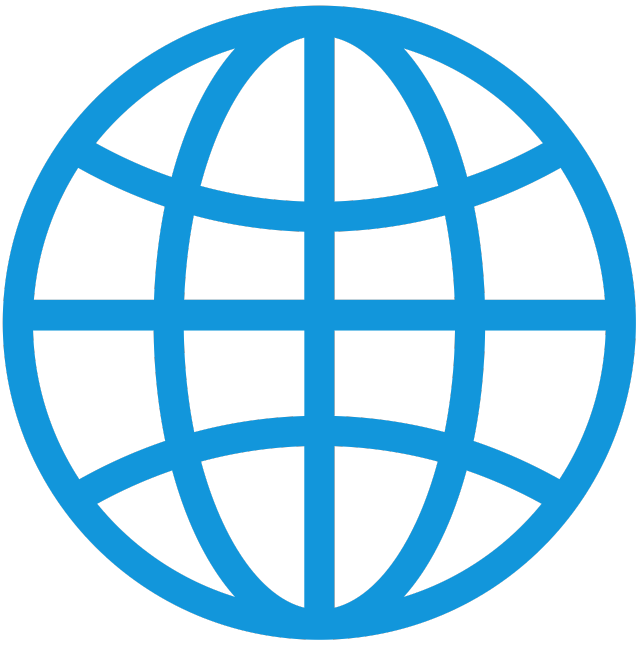}\ Homepage}: \texttt{\href{https://aopolin-lv.github.io/F1-VLA/}{\textcolor{magenta}{https://aopolin-lv.github.io/F1-VLA}}} \\
\textbf{\includegraphics[height=1.2\fontcharht\font`X]{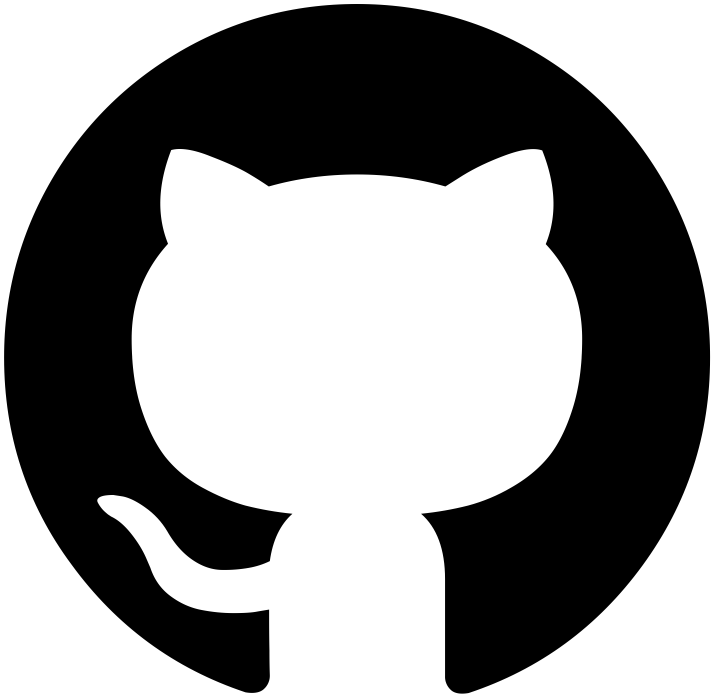}\ Github}: \texttt{\href{https://github.com/InternRobotics/F1-VLA}{\textcolor{magenta}{https://github.com/InternRobotics/F1-VLA}}}\\
\textbf{\includegraphics[height=1.2\fontcharht\font`X]{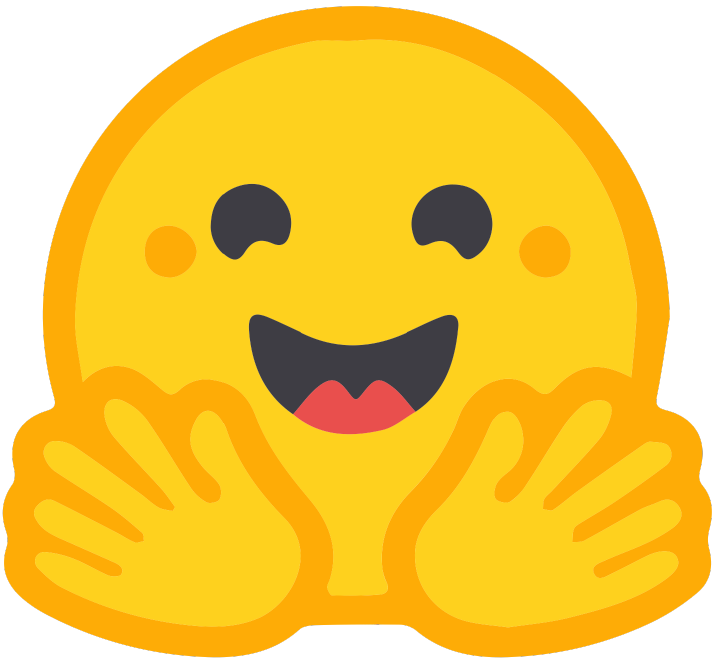}\ Huggingface}: \texttt{\href{https://huggingface.co/InternRobotics/F1-VLA}{\textcolor{magenta}{https://huggingface.co/InternRobotics/F1-VLA}}}
}

\iclrfinalcopy
\begin{document}

\maketitle

\begin{abstract}
    \input{sections/abstract}
\end{abstract}

\input{sections/introduction}
\input{sections/method}
\input{sections/experiment}
\input{sections/relatedwork}
\input{sections/conclusion}

\bibliography{iclr2026_conference}
\bibliographystyle{iclr2026_conference}

\appendix
\input{sections/appendix}

\end{document}

%% file: math_commands.tex
\usepackage{amsmath,amsfonts,bm}

\def\eqref#1{equation~\ref{#1}}
\def\1{\bm{1}}

\DeclareMathAlphabet{\mathsfit}{\encodingdefault}{\sfdefault}{m}{sl}
\SetMathAlphabet{\mathsfit}{bold}{\encodingdefault}{\sfdefault}{bx}{n}

%% file: sections/abstract.tex
Executing language-conditioned tasks in dynamic visual environments remains a central challenge in embodied AI. 
Existing Vision-Language-Action (VLA) models predominantly adopt reactive state-to-action mappings, often leading to short-sighted behaviors and poor robustness in dynamic scenes.
In this paper, we introduce \modelname, a pretrained VLA framework which integrates the \textit{visual foresight generation} into decision-making pipeline.
\modelname adopts a Mixture-of-Transformer architecture with dedicated modules for perception, foresight generation, and control, thereby bridging understanding, generation, and actions.
At its core, \modelname employs a next-scale prediction mechanism to synthesize goal-conditioned visual foresight as explicit planning targets. 
By forecasting plausible future visual states, \modelname reformulates action generation as a foresight-guided inverse dynamics problem, enabling actions that implicitly achieve visual goals.
To endow \modelname with robust and generalizable capabilities, we propose a three-stage training recipe on an extensive dataset comprising over 330k trajectories across 136 diverse tasks.
This training scheme enhances modular reasoning and equips the model with transferable visual foresight, which is critical for complex and dynamic environments.
Extensive evaluations on real-world tasks and simulation benchmarks demonstrate \modelname consistently outperforms existing approaches, achieving substantial gains in both task success rate and generalization ability.

%% file: sections/introduction.tex
\section{Introduction}
\label{sec:intro}
Vision-Language-Action (VLA) models~\citep{kim2024openvlaopensourcevisionlanguageactionmodel,team2025gemini,black2024pi_0} aim to equip robots with the ability to execute natural language instructions in visually rich environments. 
By aligning language instructions with perceptual inputs and mapping them to actions, such models enable language-guide manipulation and versatile human–robot interaction.
% However, achieving reliable VLA performance in realistic settings remains difficult, as agents must reason over diverse objects, ambiguous instructions, and dynamic scenes while producing temporally coherent control.
% These challenges expose the limitations of purely reactive action prediction and highlight the need for models that incorporate predictive foresight into decision making.
However, reliable performance in realistic settings remains elusive: environments are inherently dynamic, i.e., objects move, contexts shift, and instructions unfold over time, so robots must ground ambiguous language, handle diverse objects, and maintain long-horizon temporal coherence as scenes evolve. 
These conditions expose a core limitation of purely reactive state-to-action mappings: without predictive foresight about likely future states, policies become short-sighted and brittle under distribution shifts. 
% This motivates VLA models that integrate explicit anticipation into decision making.

\begin{figure}[tp]
    \centering
    \includegraphics[width=\linewidth]{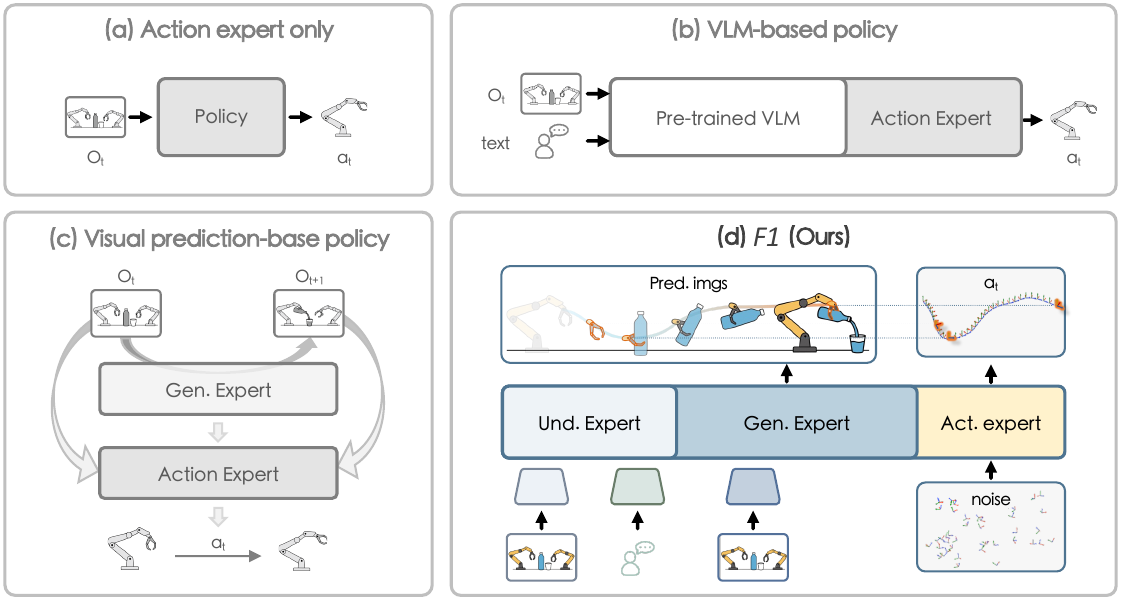}
    \caption{
        \textbf{The comparison of varied paradigms for manipulation policies.} 
        The earliest end-to-end manipulation policies are illustrated in \cref{fig:teaser}(a), such as ACT~\citep{zhao2023act} and DP~\citep{chi2023dp}. 
        Policies shown in \cref{fig:teaser}(b) introduce VLM, aiming to empower the action expert with the general capabilities of VLM in scene and instruction understanding, like $\pi_{0}$~\citep{black2024pi_0} and gr00t-N1~\citep{bjorck2025gr00t_n1}. There are also approaches, as seen in \cref{fig:teaser}(c), e.g., VPP~\citep{hu2024vpp} and Genie Envisioner~\citep{liao2025genie_envisioner}, that leverage video diffusion models to guide action execution through video prediction. 
        As depicted in \cref{fig:teaser}(d), we adopts an integrated architecture of understanding, generation, and execution, empowering the action execution module with capabilities in both scene and instruction comprehension as well as dynamic temporal prediction.
    }
    \label{fig:teaser}
    \vspace{-0.5em}
\end{figure}

Previous efforts on manipulation policy learning can be broadly grouped into three paradigms, as illustrated in \cref{fig:teaser}.
The earliest line of work employs only an \emph{action expert} trained end-to-end from observations to low-level actions~\citep{zhao2023act, chi2023dp}, but such purely reactive mappings lack semantic grounding and generalization across tasks and embodiments (\cref{fig:teaser}(a)). 
To overcome these limitations, subsequent approaches integrate Vision-Language-Models (VLMs) into the policy, leveraging pretrained multimodal knowledge to enhance scene and instruction understanding~\citep{black2024pi_0, bjorck2025gr00t_n1} (\cref{fig:teaser}(b)).
However, they lack temporal evolutionary modeling and remain reactive in nature, thus failing to cope reliably with dynamic or long-horizon manipulation tasks.
More recently, visual prediction–based policies~\citep{hu2024vpp,liao2025genie_envisioner} attempt to anticipate future observations as auxiliary signals (\cref{fig:teaser}(c)), but without integrating semantic understanding from VLMs, their predictions lack semantic grounding and result in brittle control, offering limited robustness and generalization. 

Across these paradigms, the main limitation is their reliance on reactive state-to-action mappings, which leads to short-sighted behavior and fragility under dynamic and complex manipulation tasks. 
This motivates a central question: \textbf{\textit{What architectural and training principles are required to move beyond reactive imitation and toward robust, foresight-driven policy?}}

Inspired by the Predictive Inverse Dynamics Models~\citep{tian2024predictiveinversedynamicsmodels, black2023zeroshotroboticmanipulationpretrained, du2023learninguniversalpoliciestextguided}, we introduce \modelname, a VLA framework that integrates goal-conditioned visual foresight into the perception–action loop~(\cref{fig:teaser}d). 
PIDMs predict a future state first, and then frame control as inferring the actions required to realize a desired future observation. 
By adopting this principle, \modelname reformulates action generation as foresight-guided inverse dynamics: actions are derived not only from the current observation, but also from an anticipated visual outcome. 
Specifically, \modelname adopts a Mixture-of-Transformer~(MoT)~\citep{liang2025mot} architecture with three dedicated experts for understanding, foresight generation, and action execution, thereby bridging perception, prediction, and control in a unified framework. 
To equip the model with robust and transferable capabilities, we design a progressive three-stage training recipe.

\textit{From the perspective of model architecture}, \modelname introduces two key designs. 
First, a progressive attention scheme regulates information exchange across experts: intra-expert attention captures rich token interactions, while inter-expert attention enforces a causal flow from understanding to foresight to action, ensuring stability and interpretability. 
Second, the generation expert employs a next-scale prediction mechanism~\citep{tian2024visualautoregressivemodelingscalable} to synthesize goal-conditioned visual foresight efficiently, producing explicit planning targets that guide subsequent control. 
These designs enable \modelname to explicitly couple semantic understanding, predictive foresight, and action execution within a unified backbone. 

\textit{From the perspective of training strategy}, we propose a three-stage recipe that progressively aligns and integrates the experts. 
Stage I injects foresight capability by aligning the generation expert with the understanding expert which inherits from a pretrained MLLM. 
Stage II pretrains the entire model on large-scale public robot datasets to learn general shared visuomotor knowledge. 
Stage III post-trains on task-specific data to adapt the model to new embodiments and fine-grained manipulation skills. 
This progressive scheme not only stabilizes optimization but also endows the model with generalizable foresight, which is critical for robustness in dynamic and long-horizon tasks.

Overall, we present \modelname, a pretrained Vision-Language-Action model with 4.2B parameters.
Building on predictive inverse dynamics modeling, \modelname introduces visual foresight as an explicit planning signal and couples it with action generation through a progressive training scheme. 
Extensive experiments in both real-world and simulation benchmarks demonstrate that \modelname substantially improves success rates and robustness compared to reactive baselines, particularly in dynamic and long-horizon tasks. 
In summary, our contributions are as follows:
\begin{itemize}[noitemsep,leftmargin=*]
    \item We introduce a novel paradigm for VLA models by integrating a dedicated generation expert that leverages a predictive inverse dynamics model to forecast visual observation, fundamentally transforming action prediction from a reactive to a planning-based process.
    \item To ensure robustness and generalizability, we develop a three-stage training recipe. This carefully designed scheme progressively integrates the model's distinct understanding, generation, and action modules.
    \item Our model \modelname demonstrates superiority in both simulation and real-world tasks. We show substantial improvements over reactive baselines, particularly in challenging dynamic environments and long-horizon tasks that require robust planning and foresight.
\end{itemize}

%% file: sections/method.tex
\section{The \texorpdfstring{\modelname}{F1} Framework: Bridging Perception, Foresight, to Actions}
\label{sec:method}
We present \modelname, a Vision-Language-Action (VLA) model that bridges perception, visual foresight generation, to action execution through a Mixture-of-Transformer (MoT) architecture~\citep{liang2025mot}. 
By jointly modeling understanding, foresight, and control, \modelname enables language-conditioned agents to plan and act robustly in complex environments.
\begin{figure}[t]
    \centering
    \includegraphics[width=\textwidth]{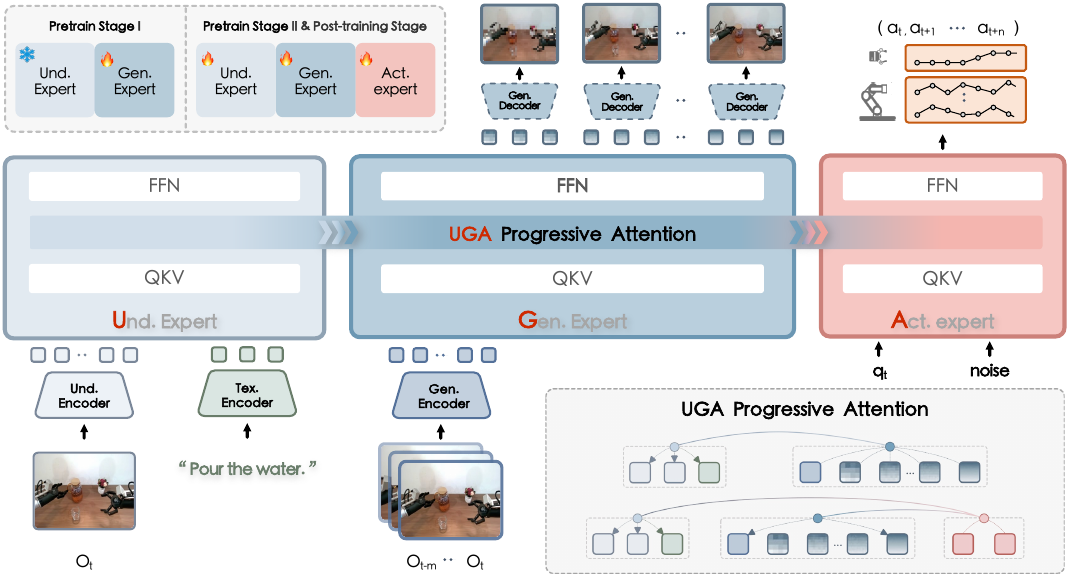} 
    \caption{
        \textbf{Overview of \modelname framework}. It employs the Mixture-of-Transformer~\citep{liang2025mot} architecture comprising three core components: an understanding expert, a generation expert, and an action expert. 
        The understanding expert processes instructions and observations to generate a foresight image. 
        The visual foresight is then fed to action expert to predict a target action via the predictive inverse dynamics modeling~\citep{tian2024predictiveinversedynamicsmodels, du2023learninguniversalpoliciestextguided}.
    }
    \vspace{-0.5em}
\label{figure:framework}
\end{figure}

\subsection{Architecture Overview}
As illustrated in~\cref{figure:framework}, \modelname comprises three dedicated experts: \textbf{an understanding expert, a generation expert, and an action expert}.
Given an instruction $l$ and current observation $o_{t}$, the understanding expert encodes semantic and visual information to establish a shared multimodal representation. 
This representation is then passed to the foresight generation expert, which predicts a goal-conditioned visual foresight $\hat{o}_{t+1}$. 
To capture temporal dynamics, the foresight module additionally leverages a sequence of past observations $\{o_{t-m}, \ldots, o_{t-1}\}$, thereby grounding the prediction in both historical context and task goals. 
Finally, the predicted foresight image $\hat{o}_{t+1}$ is fed into the action expert, which formulates a predictive inverse dynamics modeling problem, enabling the model to generate an action chunk $\hat{a}_{t:t+k}$ that drives the robot toward the synthesized visual target.

\subsection{Unified Understanding-Generation-Action Transformer}
Building on the demonstrated effectiveness of decoder-only transformers across large models~\citep{openai2024gpt4technicalreport, yang2025qwen3, bai2025qwen25vl, liu2023visualinstructiontuning}, \modelname instantiates all three experts with a common decoder-only architectural backbone, enabling scalable autoregressive modeling while retaining expert-specific specialization.

\noindent\textbf{Understanding Expert.} 
To achieve robust alignment between natural language instructions and perceptual inputs, the understanding expert is initialized from a pretrained vision–language model trained on large-scale paired text–image data. 
At each timestep, the current visual observation $o_t$ is first encoded by a SigLIP vision encoder~\citep{zhai2023sigmoidlosslanguageimage} to produce high-level perceptual features. 
These features are fused with the language prompt and processed by the decoder-only transformer, allowing the expert to capture semantic correspondences between task goals and the observed scene. 
This design equips the understanding expert with reliable, semantically aligned representations that serve as the foundation for subsequent foresight generation and action execution.

\begin{wrapfigure}{hr}{0.5\columnwidth}
    \centering
    \vspace{-4.3mm}
    \begin{minipage}[t]{0.5\columnwidth}
        \centering
        \includegraphics[width=\linewidth]{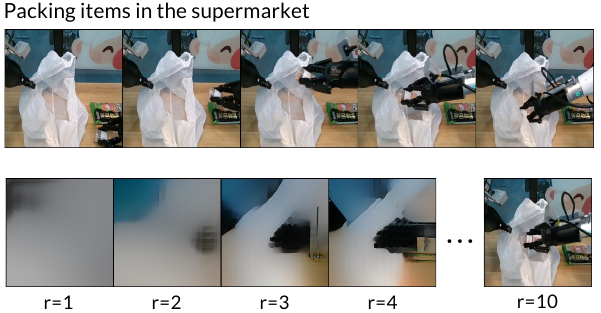}
        \vspace{-1.75em}
        \caption{Visualization of Residual VQ-VAE from 16$\times$16 to 256$\times$256 resolution.}
        \label{fig:rq_vae}
    \end{minipage}
    \vspace{-3.8mm}
\end{wrapfigure}
\noindent\textbf{Generation Expert.} 
The generation expert is designed to produce a foresight image $\hat{o}_{t+1}$ conditioned on the current observation $o_t$ and the language goal $l$, serving as an explicit intermediate target for subsequent control. 
In contrast to conventional reactive policies, this module anticipates plausible future visual states, thereby enabling smoother and more adaptive behaviors in dynamic environments. 
Efficient foresight prediction is challenging, since high-fidelity visual synthesis often incurs substantial computational cost. 
To mitigate this issue, we employ a next-scale prediction strategy that balances computational efficiency with predictive accuracy. 

Concretely, recent observations $\{o_{t-m}, \ldots, o_{t}\}$ are first encoded by a multi-scale residual vector quantization (VQ) encoder~\citep{lee2022autoregressiveimagegenerationusing}. 
As illustrated in~\cref{fig:rq_vae}, each frame is decomposed into $16\times16$ patches across $k$ spatial scales $\{r_1,\ldots,r_k\}$, yielding discrete tokens $\{z_i^0, \ldots, z_i^k\}$ for each $o_i$. 
To avoid prohibitively long sequences from concatenating tokens across multiple frames, a temporal convolutional network aggregates motion-relevant features into a compact representation. 
This representation is then processed by a decoder-only transformer to autoregressively generate foresight tokens, which are subsequently decoded into the predicted future image $\hat{o}_{t+1}$. 
Through this design, the generation expert realizes an efficient mechanism for real-time foresight, supplying explicit visual targets that guide the inverse dynamics model in action execution.

\noindent\textbf{Action Expert.} 
The action expert is responsible for mapping the multimodal context into executable robot actions. 
Conditioned on the language goal $l$, the current observation $o_t$, and the generated foresight image $\hat{o}_{t+1}$, it predicts a short-horizon action sequence $\hat{a}_{t:t+k}$. 
By incorporating foresight explicitly, the policy grounds its decisions not only in the present state but also in an anticipated visual target, which supports goal-directed and temporally consistent behavior. 
In practice, we employ a chunked action prediction~\citep{zhao2023act} which captures motion patterns over multiple steps and a flow-matching objective in continuous action space.
% rather than predicting frame-wise actions independently. 
% This design mitigates error accumulation and aligns more naturally with the continuous control dynamics of embodied agents. 
% For training, we adopt a flow-matching objective in continuous action space.
% , which enables distribution-level supervision and provides richer gradient signals than cross-entropy loss on discretized tokens. 
Through this formulation, the action expert produces accurate and coherent action plans that are responsive to immediate observations while remaining consistent with long-term task objectives.

\noindent\textbf{Attention Mechanism.} 
To coordinate these heterogeneous experts, we introduce a hierarchical scheme termed \textit{Understanding–Generation–Action (UGA) progressive attention}. 
Within each expert, bidirectional intra-expert attention enables comprehensive token interactions, except in the generation expert, where foresight tokens follow a causal, scale-conditioned pattern to preserve autoregressive consistency. 
Across experts, inter-expert attention follows a causal hierarchy: the generation expert attends to the understanding expert, while the action expert attends to both, but no information flows in the reverse direction. 
This progressive design enforces foresight as an explicit intermediate representation—preventing information leakage from actions back into foresight—thereby stabilizing training, enhancing interpretability, and ensuring that downstream control is genuinely guided by predicted visual outcomes rather than by shortcut correlations.

\subsection{Training Recipe}
We train \modelname on a mixture of open-source datasets and task-specific collections, covering more than 320k trajectories across 136 tasks and 5 embodiments. 
Comprehensive statistics of the training corpus are provided in the Appendix~\ref{appendix:data_details}. 
Our training recipe follows a three-stage paradigm designed to progressively build alignment, generalization, and task adaptation: (i) \textbf{Pretrain Stage I.}, aligning the generation expert with the understanding expert, (ii) \textbf{Pretrain Stage II}, pretraining the full model on large-scale public robot datasets, and (iii) \textbf{Post-train Stage}, post-training on task-specific demonstrations for embodiment adaptation.

\noindent\textbf{Pretrain Stage I.}
The understanding expert inherits the weight from $\pi_0$~\citep{black2024pi_0}, while the generation expert is randomly initialized. 
We then train the generation expert to synthesize future visual tokens conditioned on historical observations and language instructions, with outputs aligned to the semantic space established by the pretrained understanding expert. 
This stage injects generative foresight into the model while retaining the pretrained vision–language alignment.

Formally, given a sequence of historical observations $\{o_{t-m}, \ldots, o_t\}$ and an instruction $l$, the generation expert predicts a foresight image $\hat{o}_{t+1}$ whose VQ token representation is matched to the target. 
The training objective minimizes the negative log-likelihood of the ground-truth tokens:
\begin{equation}
\mathcal{L}_{\mathrm{gen}}^{\mathrm{gt}} = - \mathbb{E}_{\{o_i, l\} \sim \mathcal{D}} \left[ \sum_{j=1}^{N} \log p_\theta(z_j \mid {z}_{1:j-1}^{gt}, \{o_i\}_{i=t-m}^{t}, l) \right],
\end{equation}
where $z^{\mathrm{gt}}_j = E(o_{t+1})_j$ are discrete tokens encoded by the residual VQ-VAE. 
We apply teacher forcing to stabilize autoregressive training at this stage.

\noindent\textbf{Pretrain Stage II \& Post-train Stage.}
After Stage I aligns the generation and understanding experts, we jointly optimize all three experts under a unified framework. 
There are two steps: Stage II pretrains the full model on large-scale public robot datasets for foundational visuomotor learning, and the post-train stage finetunes on task-specific demonstrations for embodiment adaptation. 
There are two objectives driving this process:

(1) \textit{Autoregressive Next-Scale Prediction.} 
In contrast to the teacher-forcing regime in Stage I, we adopt an autoregressive formulation where each foresight token is generated conditioned on previously predicted tokens. 
Given historical observations $\{o_{t-m}, \ldots, o_t\}$ and language instruction $l$, the model autoregressively generates future VQ tokens $\hat{z}_{1:N}$, where each prediction step conditions on previously generated tokens:
\begin{equation}
  \mathcal{L}_{\mathrm{gen}}^{\mathrm{pred}} = - \mathbb{E}_{\{o_i, l\} \sim \mathcal{D}} \left[ \sum_{j=1}^{N} \log p_\theta(z_j \mid \hat{z}_{1:j-1}, \{o_i\}_{i=t-m}^{t}, l) \right],
\end{equation}
This autoregressive formulation enforces distributional consistency between training and inference, thereby enhancing generation stability and long-horizon coherence.

(2) \textit{Flow Matching-based Action Prediction.}
To establish a principled connection between foresight and motor control, we adopt flow matching~\citep{lipman2023flowmatchinggenerativemodeling} to model the continuous transformation from Gaussian noise to expert actions. 
Given interpolated actions $a_t^\tau = (1 - \tau)\epsilon + \tau a_t$ with $\tau \sim \mathcal{U}(0,1)$ and $\epsilon \sim \mathcal{N}(0,I)$, the policy $\pi_\theta$ learns vector fields that guide the transformation toward target actions:  
\begin{equation}
  \mathcal{L}_{\mathrm{action}} = \mathbb{E}_{\{a_t, o_i, q_t, l\} \sim \mathcal{D}} \left[ \left\| \pi_\theta(l, \{o_i\}_{i=t-m}^{t}, q_t, a_t^\tau) - (a_t - \epsilon) \right\|^2 \right],
\end{equation}
where $q_t$ denotes the proprioception information at time $t$.
% During inference, actions are synthesized via numerical integration along the learned vector field using the Euler method, effectively sampling from the learned policy distribution.

The overall training objective is defined as a weighted sum of the two components:
\begin{equation}
  \mathcal{L}_{\mathrm{total}} = \mathcal{L}_{\mathrm{gen}}^{\mathrm{pred}} + \lambda \cdot \mathcal{L}_{\mathrm{action}},
\end{equation}
with $\lambda$ balancing the objectives. 
This joint optimization enforces representational consistency across experts, integrates foresight with control, and facilitates generalization over tasks and embodiments.

\subsection{Implementation Details}
\textbf{Model Architecture.} 
\modelname adopts a Mixture-of-Transformer architecture comprising an understanding expert, a generation expert, and an action expert. 
The architecture of understanding expert is implemented the same as PaliGemma~\citep{beyer2024paligemmaversatile3bvlm}, while the generation and action experts follow the same Gemma backbone~\citep{gemmateam2025gemma3technicalreport}. 
The backbone integrates Swish activations~\citep{ramachandran2017searchingactivationfunctions}, RMSNorm normalization~\citep{zhang2019rootmeansquarelayer}, and Rotary Position Embeddings~\citep{su2023roformerenhancedtransformerrotary}. 
For initialization, the understanding and action experts are inherited from $\pi_0$~\citep{black2024pi_0}, whereas the generation expert is randomly initialized and equipped with a pretrained residual VQ-VAE from VAR for image quantization~\citep{tian2024var}.

\noindent\textbf{Dataset.} 
\modelname is trained on a large-scale corpus of robotic manipulation trajectories comprising approximately 330k episodes across 136 tasks. 
The corpus integrates widely used public benchmarks, including LIBERO~\citep{liu2023liberobenchmarkingknowledgetransfer}, Open-X-Embodiment~\citep{embodimentcollaboration2025openxembodimentroboticlearning}, and AgiBotWorld~\citep{agibotworldcontributors2025agibotworldcolosseolargescale} (see Appendix~\ref{appendix:data_details} for details). 
It spans a broad spectrum of skills, from elementary pick-and-place to complex behaviors such as grasping, handover, and pushing, and covers diverse temporal scales with episode lengths ranging from 10 seconds to over 2 minutes. 
This diversity in task complexity and temporal horizon provides rich supervision for developing robust visuomotor policies.

\noindent\textbf{Training and Inference.} 
Training proceeds in three sequential stages. 
In Pretrain Stage I, the understanding expert is frozen while the generation expert is trained with VAR to predict future images across 10 resolutions. 
In Stage II and the post-training stage, all experts are optimized jointly for end-to-end visuomotor learning. 
Further hyperparameter details, such as batch size and training steps, are provided in the Appendix~\ref{appendix:training_details}. 
For efficient real-time control during inference, foresight prediction is restricted to 4 scales.

%% file: sections/experiment.tex
\begin{figure}[h]
    \centering
    \includegraphics[width=1\linewidth]{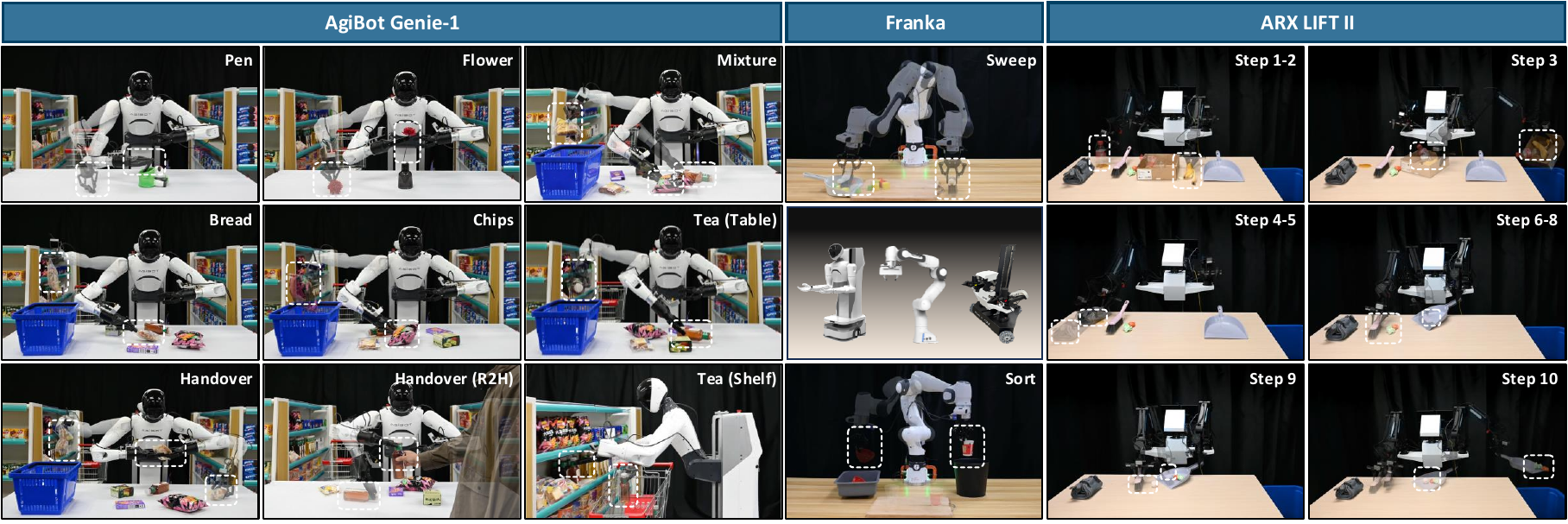}
    \caption{
        \textbf{Overview of real-world robot experiments}. 
        We conduct 12 real-world robotic experiments on three different platforms: Genie-1, Franka, and ARX LIFT II. 
        The experiments on Genie-1 are designed to evaluate the model's ability to handle task diversity. 
        The Franka experiments assess the model’s quick adaptation capabilities, while the ARX LIFT II tasks are used to benchmark its performance on challenging long-horizon manipulation problems.
    }
    \label{fig:realworld_task}
\end{figure}

\section{Experiments}
\label{sec:exp}
To thoroughly evaluate our proposed \modelname, we conduct extensive experiments across both simulated benchmarks and real-world tasks. Our evaluation not only validates the model's core performance but also investigates its robustness and generalization in various challenging scenarios.

We first present a quantitative comparison of \modelname with existing mainstream Vision-Language-Action models~\citep{black2024pi_0, kim2024openvlaopensourcevisionlanguageactionmodel, bjorck2025gr00t_n1} to demonstrate its superior performance. 
We then perform a series of comprehensive ablation studies to justify the necessity and contribution of each key component.
Furthermore, to provide deeper insights into the model’s robustness, generalization, and rapid adaptation, we conduct several additional experiments. 
Specifically, we set up a kitchen environment with a conveyor belt to evaluate its performance on a challenging dynamic manipulation task involving a moving target. 
We also conduct experiments on various robot embodiments to evaluate the model's ability to rapidly adapt to new platforms.
Finally, we evaluate its long-term robustness in complex, sequential scenarios through a long-horizon task.

\subsection{Comparison on the Real-world Tasks}
To further validate the performance of our proposed \modelname, we conduct experiments on 9 real-world tasks using the Genie robot, a dual-arm manipulation platform as illustrated in~\cref{fig:realworld_task}. 
Details of the hardware platform and task setup are provided in Appendix~\ref{appendix:task_details}. 
For each task, we collect demonstrations and fine-tune \modelname and baseline VLA models, including $\pi_0$~\citep{black2024pi_0}, gr00t-N1~\citep{bjorck2025gr00t_n1}, and gr00t-N1.5~\citep{bjorck2025gr00t_n1}, for comparison. 
Each model is evaluated 15 times per task, and we report the average grasp and task success rates.

As shown in~\cref{tab:exp_realworld}, \modelname demonstrates superior performance across all tasks, achieving an average grasp rate of 92.6\% and an average success rate of 82.2\%. 
In contrast, the best-performing baseline, $\pi_0$, only achieves 78.5\% grasp and 65.2\% success rates, while gr00t-N1 and gr00t-N1.5 perform even worse. 
These results clearly highlight the effectiveness of our foresight-guided design and modular reasoning capabilities.
In terms of individual tasks, \modelname achieves a perfect grasp rate (100\%) on some tasks with correspondingly high success rates, e.g. \textit{Pen}, \textit{Flower}, \textit{Chip}, and \textit{Tea (Table)}. 
Notably, on more challenging tasks, i.e., \textit{Handover (R2H)}, which require precise coordination and dynamic adjustment, \modelname consistently outperforms baselines by large margins, achieving up to 93.3\% success rate compared to $\pi_0$ (40\%) and gr00t-N1 (13.3\%).
The superior performance of \modelname can be attributed to its ability to predict plausible future visual states via next-scale prediction and leverage them as explicit planning targets. 
By transforming action generation into a foresight-guided inverse dynamics problem, \modelname effectively bridges understanding, generation, and action, leading to improved robustness and generalization across diverse real-world tasks.

\begin{table}[h]
 \centering
 \setlength{\tabcolsep}{4pt}
 \renewcommand{\arraystretch}{1.05}
 \resizebox{\linewidth}{!}{
    \begin{tabular}{l *{10}{>{\centering\arraybackslash}m{1.7cm}}}
    \toprule[0.3mm]
      & \textbf{Pen} & \textbf{Flower} & \textbf{Chip} & \textbf{Tea (Table)} & \textbf{Tea (Shelf)} & \textbf{Bread} & \textbf{Handover} & \textbf{Handover (R2H)} & \textbf{Mixture} & \textbf{Average} \\
    \cmidrule[0.15mm]{1-11}
        $\pi_0$      & 66.7\% & 66.7\% & 86.7\% & 86.7\% & 73.3\% & 66.7\% & 33.3\% & 40.0\% & 66.7\% & 65.2\% \\
        gr00t-N1     & 46.7\% & 33.3\% & 33.3\% & 40.0\% & 13.3\% & 33.3\% & 26.7\% & 13.3\% & 33.3\% & 30.4\% \\
        gr00t-N1. 5  & 73.3\% & 40.0\% & 46.6\% & 73.5\% & 26.6\% & 53.3\% & 60.0\% & 40.0\% & 66.7\% & 53.3\% \\
    \rowcolor{myred!20!white}
        $\mathcal{F}_1$ (Ours) & \textbf{93.3\%} & \textbf{80.0\%} & \textbf{100.0\%} & \textbf{93.3\%} & \textbf{86.7\%} & \textbf{66.7\%} & \textbf{80.0\%} & \textbf{73.3\%} & \textbf{66.7\%} & \textbf{82.2\%} \\
    \bottomrule[0.3mm]
    \end{tabular}
  }
 \caption{
  \textbf{Results on Real-world Tasks.}
  We test each model 15 times per task and report its average success ratio. The best results are denoted in \textbf{bold}.
  The experimental results show the superior performance of our \modelname comparing other VLA models, especially in the Handover (R2H) task which requires the dynamic environment understanding capabilities.
 }
 \label{tab:exp_realworld}
\end{table}

\subsection{Comparison on the Simulation Benchmark}
We evaluate \modelname against various VLA models on two simulation benchmarks: LIBERO~\citep{liu2023liberobenchmarkingknowledgetransfer} and SimplerEnv Bridge~\citep{li2024evaluatingrealworldrobotmanipulation}. 
This comparison specifically aims to demonstrate how \modelname's integrated understanding-generation-action framework, leveraging explicit visual foresight as targets, leads to superior and more robust performance compared to prevailing methods.

\noindent\textbf{LIBERO Benchmark.}
It evaluates robotic manipulation skills, focusing on spatial and object-centric reasoning, and long-horizon planning. 
We compare \modelname with Diffusion Policy~\citep{chi2023dp}, OpenVLA~\citep{kim2024openvlaopensourcevisionlanguageactionmodel}, SpatialVLA~\citep{qu2025spatialvlaexploringspatialrepresentations}, $\pi_0$~\citep{black2024pi_0}, $\pi_0$-Fast~\citep{pertsch2025fastefficientactiontokenization}, gr00t-N1~\citep{bjorck2025gr00t_n1}, and CoT-VLA~\citep{zhao2025cotvlavisualchainofthoughtreasoning}, using Success Rate (SR) and Ranking (Rank) as primary metrics. 
As shown in~\cref{tab:exp_libero}, our proposed model \modelname consistently achieves superior performance across all suites. 
This benchmark includes tasks demanding precise spatial understanding, intricate object interaction, goal-oriented execution, and particularly challenging long-horizon sequences. 
The significant improvement underscores the efficacy of \modelname's unique foresight-guided planning approach, which allows it to transform action generation into a more informed inverse dynamics problem by predicting plausible future visual states.
Notably, on the LIBERO-Long tasks, our model exhibits a pronounced advantage over reactive methods.
It highlights \modelname's enhanced capability for long-term planning and execution, a direct benefit of its explicit visual foresight. 
\begin{table}[h]
 \centering
 \renewcommand{\arraystretch}{1.1}
 \resizebox{\linewidth}{!}{
  \begin{tabular}{lccc|cc|cc|cc|cc}
  \toprule[0.45mm]
  \multirow{2}[4]{*}{}
   & \multirow{2}[4]{*}{\textbf{Pretrained}} 
   & \multicolumn{2}{c|}{\textbf{LIBERO-Spatial}} 
   & \multicolumn{2}{c|}{\textbf{LIBERO-Object}} 
   & \multicolumn{2}{c|}{\textbf{LIBERO-Goal}} 
   & \multicolumn{2}{c|}{\textbf{LIBERO-Long}} 
   & \multicolumn{2}{c}{\textbf{Average}} \\
  \cmidrule{3-12}
   & 
   & \textbf{SR ($\uparrow$)} & \textbf{Rank ($\downarrow$)} 
   & \textbf{SR ($\uparrow$)} & \textbf{Rank ($\downarrow$)} 
   & \textbf{SR ($\uparrow$)} & \textbf{Rank ($\downarrow$)} 
   & \textbf{SR ($\uparrow$)} & \textbf{Rank ($\downarrow$)} 
   & \textbf{SR ($\uparrow$)} & \textbf{Rank ($\downarrow$)} \\
  \cmidrule{1-12}
  Diffusion Policy & \textcolor{red}{\ding{55}}   & 78.5\% & 11 & 87.5\% & 10 & 73.5\% & 11 & 64.8\% & 6 & 76.1\% & 9 \\
  OpenVLA      & \textcolor{SeaGreen}{\ding{52}} & 84.7\% & 8 & 88.4\% & 9 & 79.2\% & 9 & 53.7\% & 10 & 76.5\% & 8 \\
  SpatialVLA    & \textcolor{SeaGreen}{\ding{52}} & 88.2\% & 6 & 89.9\% & 8 & 78.6\% & 10 & 55.5\% & 9 & 78.1\% & 7 \\
  $\pi_0$      & \textcolor{SeaGreen}{\ding{52}} & 98.0\% & 2 & 96.8\% & 4 & 94.4\% & 2 & 88.4\% & 3 & 94.4\% & 2 \\
  $\pi_0$-Fast   & \textcolor{SeaGreen}{\ding{52}} & 96.4\% & 3 & 96.8\% & 4 & 88.6\% & 5 & 60.2\% & 8 & 85.5\% & 5 \\
  gr00t-N1     & \textcolor{SeaGreen}{\ding{52}} & 94.4\% & 5 & 97.6\% & 1 & 93.0\% & 4 & 90.6\% & 2 & 93.9\% & 4 \\
  CoT-VLA      & \textcolor{SeaGreen}{\ding{52}} & 87.5\% & 7 & 91.6\% & 7 & 87.6\% & 6 & 69.0\% & 4 & 83.9\% & 6 \\
  \rowcolor{myred!20!white}
  \modelname (Ours) & \textcolor{red}{\ding{55}}   & 97.4\% & 3 & 97.6\% & 2 & 94.2\% & 3 & 88.0\% & 4 & 94.3\% & 3 \\
  \rowcolor{myred!20!white}
  \modelname (Ours) & \textcolor{SeaGreen}{\ding{52}} & \textbf{98.2\%} & 1 & \textbf{97.8\%} & 1 & \textbf{95.4\%} & 1 & \textbf{91.3\%} & 1 & \textbf{95.7\%} & 1 \\
  \bottomrule[0.45mm]
  \end{tabular}
 }
 \caption{
  \textbf{Results on LIBERO Benchmark.}
  We report Success Rate (SR, higher is better) and Ranking (Rank, lower is better) across four suites: LIBERO-Spatial, LIBERO-Object, LIBERO-Goal, and LIBERO-Long, as well as the averaged results. 
  The best results are displayed in \textbf{bold}.
 }
 \label{tab:exp_libero}
\end{table}

\noindent\textbf{SimplerEnv Bridge Benchmark.}
It focuses on complex, multi-step manipulation tasks that often require fine-grained control and precise interaction. 
We compare \modelname against RT-1-X~\citep{brohan2023rt1roboticstransformerrealworld}, RoboVLM~\citep{li2024generalistrobotpoliciesmatters}, SpatialVLA, $\pi_0$, and $\pi_0$-Fast, measuring Grasp Success and Overall Success for each task.
\cref{tab:exp_simpler_bridge} shows the performance of our model and other methods. 
Overall, \modelname substantially outperforms other models, demonstrating a marked improvement in average success rate compared to the next best baselines. 
This superior performance is largely attributed to \modelname's enhanced generalization capabilities for precise pick and place positions, which are critical for success in these intricate manipulation scenarios. 
We induce that unlike reactive policies that might struggle with different source object configurations or target locations, \modelname's foresight mechanism enables it to adapt more effectively. 

\begin{table}[h]
 \centering
 \scriptsize
 \renewcommand{\arraystretch}{1.1}
 \resizebox{\linewidth}{!}{
  \begin{tabular}{lccc|cc|cc|cc|cc}
    \toprule[0.65pt]
    \multirow{2}[4]{*}{}
     & \multirow{2}[4]{*}{\textbf{Pretrained}}
     & \multicolumn{2}{c|}{\textbf{Carrot on Plate}}
     & \multicolumn{2}{c|}{\textbf{Eggplant in Basket}}
     & \multicolumn{2}{c|}{\textbf{Spoon on Towel}}
     & \multicolumn{2}{c|}{\textbf{Stack Block}}
     & \textbf{Overall} \\
    \cmidrule[0.35pt]{3-11}
     &    
     & \textbf{Grasp} & \textbf{Success} 
     & \textbf{Grasp} & \textbf{Success}
     & \textbf{Grasp} & \textbf{Success}
     & \textbf{Grasp} & \textbf{Success}
     & \textbf{Average} \\
    \midrule[0.25pt]
    RT-1-X      & \textcolor{red}{\ding{55}}   & 20.8\% & 4.2\% & 0.0\% & 0.0\% & 16.7\% & 0.0\% & 8.3\% & 0.0\% & 6.3\% \\
    % Octo-Base     & \textcolor{red}{\ding{55}}   & 52.8\% & 8.3\% & 31.9\% & 0.0\% & 66.7\% & 43.1\% & 34.7\% & 12.5\% & 31.3\% \\
    % Octo-Small    & \textcolor{red}{\ding{55}}   & 27.8\% & 9.7\% & 40.3\% & 4.2\% & 87.5\% & 56.9\% & 77.8\% & 47.2\% & 43.9\% \\
    RoboVLM      & \textcolor{SeaGreen}{\ding{52}} & 25.0\% & 25.0\% & 45.8\% & 12.5\% & 58.3\% & 58.3\% & 54.2\% & 29.2\% & 38.5\% \\
    SpatialVLA    & \textcolor{SeaGreen}{\ding{52}} & 29.2\% & 25.0\% & \textbf{100.0\%} & \textbf{100.0\%} & 20.8\% & 16.7\% & 62.5\% & 29.2\% & 47.9\% \\
    $\pi_0$      & \textcolor{SeaGreen}{\ding{52}} & 25.0\% & 0.0\% & 50.0\% & 16.6\% & \textbf{91.6\%} & \textbf{62.5}\% & 45.8\% & 29.1\% & 40.1\% \\
    $\pi_0$-Fast   & \textcolor{SeaGreen}{\ding{52}} & 58.5\% & 21.9\% & 54.0\% & 10.8\% & 83.3\% & 66.6\% & 62.5\% & 29.1\% & 48.3\% \\
    \rowcolor{myred!20!white}
    \modelname (Ours) & \textcolor{red}{\ding{55}}   & 70.8\% & 33.3\% & 87.5\% & 75.0\% & 70.8\% & 45.8\% & 83.3\% & \textbf{62.5\%} & 66.1\% \\
    \rowcolor{myred!20!white}
    \modelname (Ours) & \textcolor{SeaGreen}{\ding{52}} & \textbf{87.5\%} & \textbf{70.8\%} & \textbf{100.0\%} & 66.7\% & 70.8\% & 50.0\% & \textbf{87.5\%} & 50.0\% & \textbf{72.9\%} \\
  \bottomrule[0.65pt]
  \end{tabular}
 }
 \caption{
  \textbf{Results on SimplerEnv Bridge Benchmark.}
  For each task, we report both \emph{Grasp Success} and \emph{Overall Task Success}, capturing fine-grained control and precise placement capabilities. 
  The last column shows the overall average score, obtained by averaging the success rates of grasping and full task completion across all tasks.
  The best results are displayed in \textbf{bold}. 
 }
 \label{tab:exp_simpler_bridge}
\end{table}

\subsection{Ablation Studies}
\begin{table}[tp]
 \centering
 \scriptsize
 \setlength{\tabcolsep}{4pt}
 \renewcommand{\arraystretch}{1.05}
 \resizebox{0.94\linewidth}{!}{
  \begin{tabular}{cccccccccc}
   \toprule
    & \multicolumn{4}{c}{\textbf{LIBERO}}
    & \multicolumn{4}{c}{\textbf{SimplerEnv Bridge}}
    & \multicolumn{1}{c}{\multirow{2}[2]{*}{\textbf{Avg}}}\\
   \cmidrule(r){2-5} \cmidrule(l){6-9} 
    & \multicolumn{1}{c}{\textbf{Spatial}}
    & \multicolumn{1}{c}{\textbf{Object}}
    & \multicolumn{1}{c}{\textbf{Goal}}
    & \multicolumn{1}{c}{\textbf{Long}}
    & \multicolumn{1}{c}{\textbf{Carrot}}
    & \multicolumn{1}{c}{\textbf{Eggplant}}
    & \multicolumn{1}{c}{\textbf{Spoon}}
    & \multicolumn{1}{c}{\textbf{Block}}
    & \\
   \midrule
   \rowcolor{myred!20!white}
   \multicolumn{10}{l}{\textit{Comparison of \modelname and its variant which freezes the generation expert during Stage II and III training}} \\
   \multicolumn{1}{l}{\modelname} & 98.2\% & 97.8\% & 95.4\% & 91.3\% & 70.8\% & 66.7\% & 50.0\% & 50.0\% & 77.5\% \\
   \multicolumn{1}{l}{(a) Frozen-Gen} & 95.6\% & 97.0\% & 93.6\% & 87.4\% & 41.7\% & 54.2\% & 45.8\% & 75.0\% & 73.8\% \\
   \midrule
   \rowcolor{myred!20!white}
   \multicolumn{10}{l}{\textit{Comparison of \modelname and its variant which is only trained by Stage I and Stage III}} \\
   \multicolumn{1}{l}{\modelname} & 98.2\% & 97.8\% & 95.4\% & 91.3\% & 70.8\% & 66.7\% & 50.0\% & 50.0\% & 77.5\% \\
   \multicolumn{1}{l}{(b) Cotrain-Scratch} & 97.4\% & 97.6\% & 94.2\% & 88.0\% & 33.3\% & 75.0\% & 45.8\% & 62.5\% & 74.2\% \\
   \midrule
   \rowcolor{myred!20!white}
   \multicolumn{10}{l}{\textit{Comparison of \modelname and its variant which does not have generation expert}} \\
   \multicolumn{1}{l}{\modelname} & 98.2\% & 97.8\% & 95.4\% & 91.3\% & 70.8\% & 66.7\% & 50.0\% & 50.0\% & 77.5\% \\
   \multicolumn{1}{l}{(c) No-Gen} & 98.0\% & 96.8\% & 94.4\% & 88.4\% & 0.0 \% & 16.6\% & 62.5\% & 29.1\% & 60.3\% \\
   \midrule
   \rowcolor{myred!20!white}
   \multicolumn{10}{l}{\textit{Comparison of \modelname and its variant with varying numbers of planning scales}} \\
   \multicolumn{1}{l}{\modelname (4-Scales)} & 98.2\% & 97.8\% & 95.4\% & 91.3\% & 70.8\% & 66.7\% & 50.0\% & 50.0\% & 77.5\% \\
   \multicolumn{1}{l}{(d) 2-Scales} & 98.0\% & 97.6\% & 94.6\% & 88.6\% & 29.2\% & 83.3\% & 41.7\% & 54.2\% & 73.4\% \\
   \multicolumn{1}{l}{(e) 6-Scales} & 97.0\% & 97.2\% & 94.2\% & 89.2\% & 50.0\% & 87.5\% & 37.5\% & 54.2\% & 75.8\% \\
   \bottomrule
  \end{tabular}
 }
 \caption{
  \textbf{Ablation Studies on the Simulation Benchmark.} 
  We design five model variants to verify the effects of different components of \modelname: (a) Frozen-Gen, (b) Cotrain-Scratch, (c) No-Gen, (d) 2-Scales, and (e) 6-Scales. 
  }
 \label{tab:exp_abl_sim}
\end{table}
\subsubsection{Results on the Simulation Benchmark}
To better understand the individual contributions of foresight prediction and pretraining in \modelname, we conduct systematic ablations on the LIBERO and SimplerEnv-Bridge benchmarks.
We design five model variants to isolate the effects of visual foresight and pretraining under different configurations: 
\vspace{-2em}
\begin{enumerate}[label=(\alph*),noitemsep,leftmargin=*]
    \item \textbf{Frozen-Gen}: The generation expert is pretrained in Stage I and then frozen during Pretrain Stage II and Post-train Stage III. Its predicted foresight tokens are used as planning cues to guide action prediction, but not used as training targets.
    \item \textbf{Cotrain-Scratch}: The model is trained by Pretrain Stage I and Post-train Stage III, without Pretrain Stage II, i.e., pretraining on large-scale robotic datasets.
    \item \textbf{No-Gen}: The generation expert is entirely removed, resulting in a purely VLA model without any foresight prediction or pretraining.
    \item \textbf{2-Scales}: The fully pretrained model predicts foresight tokens for only two future steps at inference time.
    \item \textbf{6-Scales}: The fully pretrained model predicts foresight tokens for six future steps at inference, evaluating the effect of different scales of intermediate visual representation.
\end{enumerate}

\noindent\textbf{Training vs. Freezing the Generation Expert.}
To understand the benefits of joint optimization, we compared our model with the (a) Frozen-Gen variant. 
In this variant, the generation expert was pretrained in Stage I but then kept frozen, preventing it from adapting during subsequent stages.
This comparison revealed a moderate performance drop from 77.5\% to 73.8\% for the Frozen-Gen variant.
Moreover, it also confirms that while a fixed, pretrained generation expert can still provide useful planning cues, end-to-end adaptation is crucial for achieving better task alignment. 
The performance gap indicates that allowing the generation expert to be fine-tuned during later stages enables it to better capture task-specific dynamics.
Ultimately, this highlights the significant benefit of jointly optimizing the foresight prediction with the downstream control policy.

\noindent\textbf{Effectiveness of Pretrain.}
We assess the impact of pretraining by comparing our model with the (b) Cotrain-Scratch variant, which was trained without the large-scale robotics dataset pretraining (Stage II). 
The removal of this stage results in an overall performance drop of approximately 3.3\%.
This finding suggests that pretraining on robotics data acts as a crucial inductive prior. 
It effectively stabilizes the entire optimization process and allows the model to inherit foundational manipulation skills. 
As a result, our model achieves a higher success rate on downstream tasks by building upon the robust capabilities acquired during the pretraining stage.

\noindent\textbf{Effectiveness of Generation Expert.}
To evaluate the contribution of the generation expert, we conduct a key ablation study by comparing our full model with the (c) No-Gen variant, where the entire visual foresight branch was removed. 
As the results in Table 1 show, this operation leads to a significant performance drop from 77.5\% to 60.3\%.
This marked degradation highlights the critical role of the generation expert in providing explicit visual foresight for effective task planning. 
The absence of foresight signals severely impairs the model's ability to achieve generalized goal alignment, reducing it to a more reactive policy. 
These findings conclusively demonstrate that the generation expert provides essential high-level guidance, enabling the policy to move beyond simple reactive behaviors and make more informed, deliberate planning decisions.

\noindent\textbf{Impact of Planning Scales.}
We assess the impact of the foresight horizon length (planning scales) by comparing our model with variants that used different numbers of future steps for foresight prediction: (d) 2-Scales, (\modelname) 4-Scales, and (e) 6-Scales. 
As shown in~\cref{tab:exp_abl_sim}, these models consistently outperform the No-Gen baseline, underscoring the effectiveness of explicit visual foresight as a planning mechanism. 
While increasing the planning scale from 2 to 6 steps generally improved performance, the 4-Scales configuration struck the optimal balance. 
This suggests that a planning horizon of four steps provides sufficient temporal abstraction to guide the policy without introducing unnecessary noise or computational overhead, thus yielding the most robust and effective results. 
This finding highlights the importance of selecting an appropriate planning scale to maximize the benefits of visual foresight.

\subsubsection{Ablation Studies on the Real-World Tasks}
To further verify the contribution of pretrain and generation which are two key components within \modelname, we conduct ablation studies on real-world tasks.
Specifically, we compare the performance of our complete model \modelname against two variants: 
\vspace{-1em}
\begin{enumerate}[noitemsep,leftmargin=*]
    \item Cotrain-Scratch: the variant of \modelname which removes the pretraining of Stage II;
    \item No-Gen: the variant which removes the foresight-related module, without any pretraining.
\end{enumerate}

\begin{figure}[h]
 \centering
 \adjustbox{width=\linewidth,center}{%
  \includegraphics{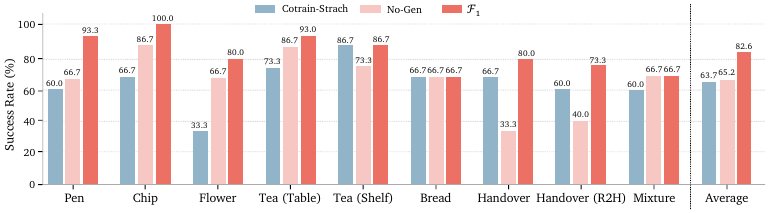}
 }
 \caption{
  \textbf{Ablation Studies on Real-world Tasks.} 
    We compare \modelname with $\pi_0$ as well as a variant that removes Pretrain Stage II. 
    For each task, we conduct 15 trials to ensure statistical reliability. 
    The results show that without Pretrain Stage II, \modelname suffers a substantial performance drop, even falling below $\pi_0$. 
    In contrast, incorporating Pretrain Stage II leads to marked improvements, with \modelname significantly surpassing the baseline.
 }
 \label{fig:exp_abl_realworld}
 % \vspace{-1mm}
\end{figure}

The results, as depicted in~\cref{fig:exp_abl_realworld}, provide clear insights into the role of each component. 
\modelname consistently outperforms both the Cotrain-Scratch variant and No-Gen across all evaluated tasks. 
A notable performance gap is observed when comparing \modelname with No-Gen, underscoring the critical role of the generation expert. 
For complex tasks such as ``Handover (R2H)'' and ``Mixture'', No-Gen only obtains 40.0\% and 60.0\% entire success rate, respectively, while our model achieves a much higher success rate of 93.3\% and 73.3\%. 
This contrast highlights that the visual foresight provided by the generation expert is essential for handling complex task dynamics and achieving robust goal alignment. 
Furthermore, the pretraining stage also proves to be a crucial component. 
When comparing \modelname with Cotrain-Scratch variant, our model shows a clear performance advantage in many tasks indicating that pretraining on a large robotics dataset provides a strong inductive prior. 
This prior enables the model to acquire foundational manipulation skills, which significantly improves generalization.

\begin{figure}[h]
 \centering
 \adjustbox{width=\linewidth,center}{
  \includegraphics{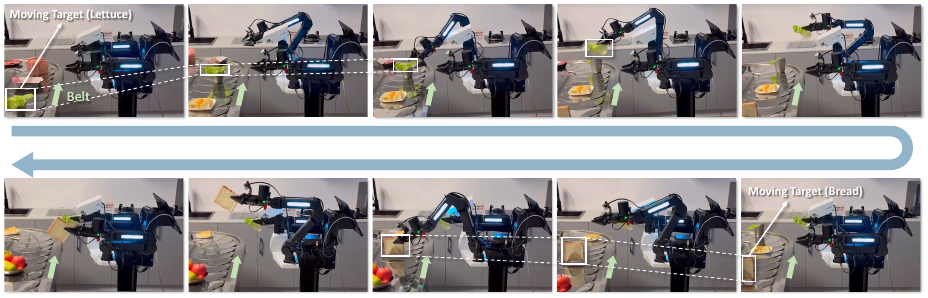}
 }
 \caption{
  \textbf{Visualization of dynamic manipulation task.}
  A kitchen environment is set up with a moving belt, where the robot must grasp a specified food item according to a language prompt while objects continuously move along the belt. 
 }
 \label{fig:exp_dyn_setting}
 \vspace{-1mm}
\end{figure}

\vspace{-2mm}
\subsection{Robustness and Generalization}
\subsubsection{Dynamic Environment}
\begin{wrapfigure}{hr}{0.3\columnwidth}
    \vspace{-4.2mm}
    \begin{minipage}[t]{0.3\columnwidth}
        \flushleft
        \includegraphics[width=\linewidth]{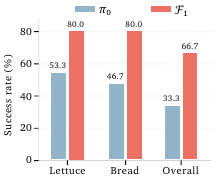}
        \caption{\textbf{Results of the Dynamic Manipulation Task.}}
        \label{fig:exp_dynamic}
    \end{minipage}
    \vspace{-7mm}
\end{wrapfigure}
To evaluate \modelname's robustness, we set up a dynamic manipulation task. As depicted in~\cref{fig:exp_dyn_setting}, we construct a kitchen environment with a moving belt. 
The robot is required to grasp a specific food item based on a given prompt, while the items are on a moving belt. 
To further test the model's generalization capabilities, we adopt a novel robot, ARX LIFT II, which is absent from our pretraining dataset. 
The number of post-train demonstrations is only 47 to explore the shared control capability obtained from pretraining stages.

As shown in Figure 3, our \modelname achieves a remarkable 66.7\% success rate for continuous dual-arm dynamic grasping. 
For both the ``Lettuce'' and ``Bread'' tasks, our model achieves an impressive 80.0\% success rate, while $\pi_0$ obtains only 53.3\% and 46.7\%, respectively.
This performance stands in stark contrast to the $\pi_0$, which achieves a success rate of only 33.3\%. 
\modelname's superior performance is directly attributable to its core visual foresight module, which enables it to predict the future position of the moving object and plan its actions accordingly. 
A detailed breakdown reveals the source of \modelname's superior performance. 
It demonstrates that \modelname effectively leverages its pretrained visual knowledge to generalize to novel embodiments and robustly handle dynamic, real-world challenges.

\vspace{-2mm}
\subsubsection{Adaptation Learning}
To further validate the generalization capabilities of our model, we conduct two additional sets of experiments, i.e., \textit{sweep} and \textit{sort}, on a Franka robotic arm. 
For the sweeping task, we evaluate three key performance metrics: the number of objects successfully swept, the maximum number of attempts required to complete the task (capped at five), and the number of empty sweeps. 
For the sorting task, we measure the success rate over three consecutive grasps.

The results are shown in~\cref{tab:exp_franka}.
In the sweeping task, our model achieves more efficient and reliable interactions, reflected in higher success rates, fewer attempts, and notably fewer empty sweeps. 
The latter are largely attributable to vertical misalignment, and their reduction indicates more precise spatial grounding during execution. 
In the sorting task, our model also outperforms $\pi_0$, particularly in the second and third consecutive grasps, where success rates increase substantially. 
This suggests that \modelname is better able to sustain performance across repeated interactions, highlighting enhanced robustness in sequential manipulation scenarios.

\begin{table}[ht]
 \centering
 \scriptsize
 \setlength{\tabcolsep}{4pt}
 \renewcommand{\arraystretch}{1.05}
 \resizebox{\linewidth}{!}{
  \begin{tabular}{cccc>{\centering\arraybackslash}p{1.3cm}>{\centering\arraybackslash}p{1.3cm}>{\centering\arraybackslash}p{1.3cm}}
   \toprule
    & \multicolumn{3}{c}{\textbf{Sweep}}
    & \multicolumn{3}{c}{\textbf{Success Rate of Consecutive Sort}} \\
   \cmidrule(r){2-4} \cmidrule(l){5-7} 
    & \multicolumn{1}{c}{\textbf{\# Succ. Objects ($\uparrow$)}}
    & \multicolumn{1}{c}{\textbf{\# Max Trials ($\downarrow$)}}
    & \multicolumn{1}{c}{\textbf{\# Empty Sweeps ($\downarrow$)}}
    & \multicolumn{1}{c}{\textbf{1st ($\uparrow$)}}
    & \multicolumn{1}{c}{\textbf{2nd ($\uparrow$)}}
    & \multicolumn{1}{c}{\textbf{3rd ($\uparrow$)}} \\
   \midrule
   $\pi_0$      & 4.9/8.0 & 4.8/5.0 & 2.4/5.0 & 100.0\% & 86.7\% & 53.3\% \\
   \rowcolor{myred!20!white}
   \modelname & 7.1/8.0 & 3.5/5.0 & 0.8/5.0 & 100.0\% & 100.0\% & 66.7\% \\
   \bottomrule
  \end{tabular}
 }
 \caption{
    \textbf{Experimental Results on Franka arm.} 
      The values for the \textit{sweep} task represent the average performance over multiple trials. For the \textit{sort} task, we report the success rate for each of three consecutive grasping attempts.
  }
 \label{tab:exp_franka}
\end{table}
\begin{table}[h]
 \centering
 \scriptsize
 \setlength{\tabcolsep}{4pt}
 \renewcommand{\arraystretch}{1.05}
 \resizebox{\linewidth}{!}{
  \begin{tabular}{cccccc}
   \toprule
    & \textbf{\ding{172} Pickplace Coke $\rightarrow$}
    & \textbf{\ding{173} Pickplace Banana $\rightarrow$}
    & \textbf{\ding{174} Pour Box $\rightarrow$}
    & \textbf{\ding{175} Pickup Cloth $\rightarrow$}
    & \textbf{\ding{176} Wipe Table $\rightarrow$}\\
   \midrule
   $\pi_0$      &  93.3\% &  93.3\% &   0.0\% &  0.0\% &  0.0\% \\
   \rowcolor{myred!20!white}
   \modelname & 100.0\% & 100.0\% & 100.0\% & 93.3\% & 93.3\% \\
   \midrule[0.08mm]
   \midrule[0.08mm]
    & \textbf{\ding{177} Pickup Broom $\rightarrow$}
    & \textbf{\ding{178} Pickup Dustpan $\rightarrow$}
    & \textbf{\ding{179} Sweeping $\rightarrow$}
    & \textbf{\ding{180} Catch Rolling Ball $\rightarrow$}
    & \textbf{\ding{181} Pour Dustpan}\\
   \midrule
   $\pi_0$      &  0.0\% &  0.0\% &  0.0\% &  0.0\% &  0.0\% \\
   \rowcolor{myred!20!white}
   \modelname & 73.3\% & 60.0\% & 40.0\% & 40.0\% & 40.0\% \\
   \bottomrule
  \end{tabular}
 }
 \caption{
    \textbf{Step-wise success rates in the long-horizon task.} 
    The task involves ten sequential steps spanning approximately two minutes. 
    Each column reports the average success rate of a specific step across 15 trials. 
    $\pi_0$ struggles beyond the first 4 stages, while \modelname achieves consistently high success rates in early steps and maintains non-trivial performance across later more complex stages.
  }
 \label{tab:exp_long}
\end{table}

\vspace{-2mm}
\subsubsection{Long-horizon Task}
To further examine the planning and foresight capabilities of \modelname, 
we design a long-horizon manipulation task on the ARX LIFT II platform. 
This task consists of ten sequential steps and requires approximately two minutes to complete. 
Unlike short episodic interactions, the long-horizon setting places greater demands on temporal consistency, error recovery, and the ability to sustain goal-directed behavior across multiple stages. 
By evaluating under this setup, we aim to assess whether the foresight-guided mechanism in \modelname can support coherent action sequences over extended durations, thereby testing its robustness in realistic, temporally extended scenarios.

The results in~\cref{tab:exp_long} reveal a stark contrast between the baseline and our model. 
The baseline policy $\pi_0$ can complete only the simplest grasping actions, but consistently fails when confronted with more complex interactions such as pouring, wiping, or multi-object coordination. 
In contrast, \modelname achieves near-perfect performance in the initial stages and sustains meaningful success rates throughout the later steps. 
Although its performance gradually decreases as the sequence progresses, this trend is expected in long-horizon scenarios due to error accumulation and the compounding difficulty of temporally extended reasoning. 
The ability of \modelname to complete the majority of the ten steps, while maintaining robustness over a two-minute execution horizon, highlights its capacity for foresight-guided planning and resilience in sequential, real-world tasks.

\section{Relationships between Generation Quality and Actions}
In \modelname, the generation expert acts as a visual planner, and the quality of its foresight images is crucial for ensuring the reliability of the action expert.
To investigate this relationship, we perform a two-step analysis.
First, we assess the foresight image quality using a set of predefined evaluation metrics.
Second, we study the correlation between generation quality and action prediction by examining token-level accuracy.

\begin{figure}[t]
 \centering
 \adjustbox{width=1.02\linewidth,center}{%
  \includegraphics{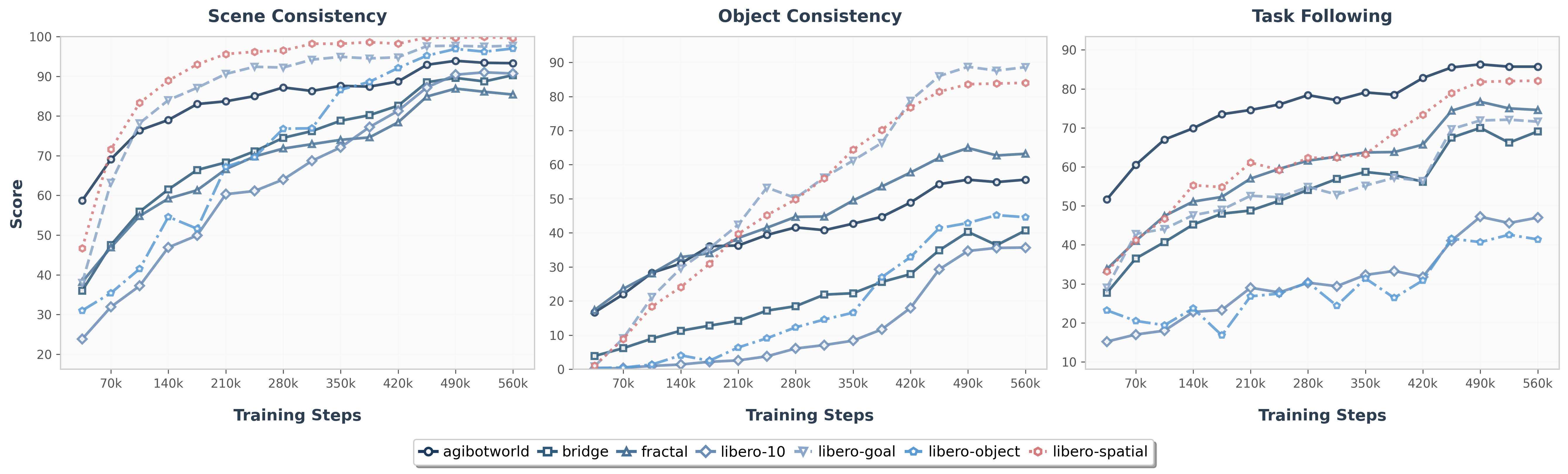}
 }
 \caption{
  \textbf{Generation Quality across Training Steps.}
    We report the evolution of generation quality along three dimensions: scene consistency (left), object consistency (middle), and task progress following (right). The x-axis denotes training steps, and different curves correspond to distinct task subsets (AgibotWorld, Bridge, Fractal, and LIBERO). Higher scores indicate a better alignment between the generated future observations and the ground-truth task progression.
 }
 \label{fig:exp_gen_quality}
 \vspace{-2mm}
\end{figure}

\subsection{Quantitative Analysis of Generation}
Unlike conventional image-generation works that emphasize pixel-level or distribution-level metrics, e.g, FID~\citep{heusel2018ganstrainedtimescaleupdate}, or PSNR, our objective is not merely to measure visual realism, but to evaluate whether generated future observations provide actionable guidance for downstream control. 
To this end, we design a multimodal evaluation protocol based on a large vision–language model, Qwen2.5-VL-32B-Instruct~\citep{bai2025qwen25vl}.

The evaluator receives a task instruction, four historical frames of the robot executing the task, one predicted next-step frame, and the ground-truth frame. 
The detailed prompt template is included in Appendix~\ref{appendix:prompt_template}.
The evaluation targets three dimensions directly related to action feasibility:
\vspace{-1em}
\begin{enumerate}[noitemsep,leftmargin=*]
    \item \textbf{Scene Consistency}: Whether the global environment remains coherent in layout, lighting, and texture, with blurry or structurally incoherent generations penalized.
    \item \textbf{Object Consistency}: Whether manipulated objects and the robot remain consistent in identity and spatial position, penalizing missing, deformed, or hallucinated objects.
    \item \textbf{Task Progress Following}: Whether the generated frame depicts a plausible next step toward fulfilling the instruction, consistent with the ground-truth trajectory.
\end{enumerate}
\vspace{-1em}
Each aspect is scored in a binary manner (0/1), and the aggregated score forms our measure of generation quality. This design explicitly ties evaluation to task relevance, enabling us to study how higher-quality generations correlate with improved action execution.
\cref{fig:exp_gen_quality} shows that the model’s visual planning capabilities develop hierarchically.
\textbf{Scene Consistency} improves rapidly at early stages, indicating that global coherence is easier to acquire. 
\textbf{Object Consistency}, however, presents a major challenge: without pretraining on large-scale object-centric datasets, the model struggles to preserve fine-grained shapes and positions, resulting in flatter curves and lower scores throughout training. 
Despite this weakness, \textbf{Task Progress Following} improves steadily and often surpasses object consistency, suggesting that the model captures high-level temporal task logic even without pixel-perfect object representation. 
This finding highlights the ability of generation expert to generate actionable foresight images, which is ultimately more critical for downstream control than exact visual fidelity.
\begin{figure}[t]
 \centering
 \adjustbox{width=1\linewidth,center}{%
  \includegraphics{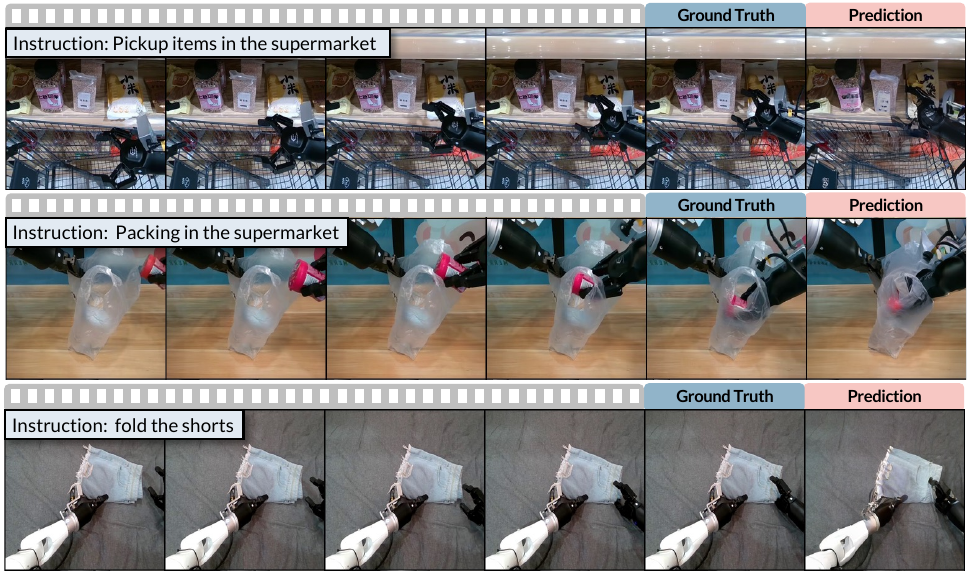}
 }
 \caption{
  \textbf{Visualization of Generated Future Images.} 
    It demonstrates the ability of \modelname to generate plausible next-step frames for various manipulation tasks, including supermarket item pickup, supermarket packing, and folding shorts. Each row compares a Ground Truth frame with a Prediction from \modelname, showcasing its accurate foresight across diverse scenarios.
 }
 \label{fig:gen_expert_visualization}
\end{figure}

\subsection{Qualitative Analysis of Generation}
In addition to the quantitative evaluation, we further conduct a qualitative analysis to gain deeper insights into the strengths and limitations of the generation expert. 
\cref{fig:gen_expert_visualization} presents representative examples of generated foresight images across diverse tasks, such as supermarket manipulation, and clothing folding.
Overall, the predicted frames capture task-level plausibility and remain aligned with the ground-truth trajectories, suggesting that the model internalizes a temporal understanding of task logic rather than merely memorizing visual appearances. 
This ability to generate plausible next-step states highlights the role of the generation expert as a visual planner.

At the same time, we observe clear limitations in visual fidelity, particularly in cases involving fine-grained object details or deformable objects, e.g., grid-shape shopping cart, plastic bags, and clothing. 
A key reason for this weakness is that our model has not been pretrained on large-scale generative datasets, making it more difficult to preserve precise object shapes and textures.
Nevertheless, these imperfections rarely prevent the predictions from providing actionable guidance for downstream control, since the generated frames still convey the essential task progression.
These qualitative observations complement our quantitative results: while pixel-level precision remains challenging, the generation expert consistently produces foresight images which are sufficiently informative to support action planning.

\subsection{Correlation between Generation and Actions}
To further study the connection between generation quality and action reliability, we perform controlled experiments on the LIBERO benchmark~\citep{liu2023liberobenchmarkingknowledgetransfer}. 
During training, our model jointly optimizes two objectives: (i) the next visual state through next-scale prediction, and (ii) the action via flow matching.
To measure progress on these two objectives, we employ accuracy metrics for both the image and action modalities.

For the image modality, we adopt a Residual VQ-VAE representation, which formulates image prediction as a token-level classification problem. 
Accordingly, we measure image token accuracy, defined as the classification accuracy of predicted visual tokens against the ground truth tokens;
For the action modality, we compute the action token accuracy via:
\begin{equation}
    \mathrm{Acc}_{\tau}=\frac{1}{N}\sum^{N}_{t=1}\left[|\hat{a}_t-a_t|<\tau\right],
\end{equation}
where $N$ is the total number of action tokens and $\tau$ denotes the error tolerance threshold. 
Since training is performed at the chunk level, we evaluate accuracy under the same granularity.

\begin{figure}[h]
 \centering
 \adjustbox{width=1\linewidth,center}{
  \includegraphics{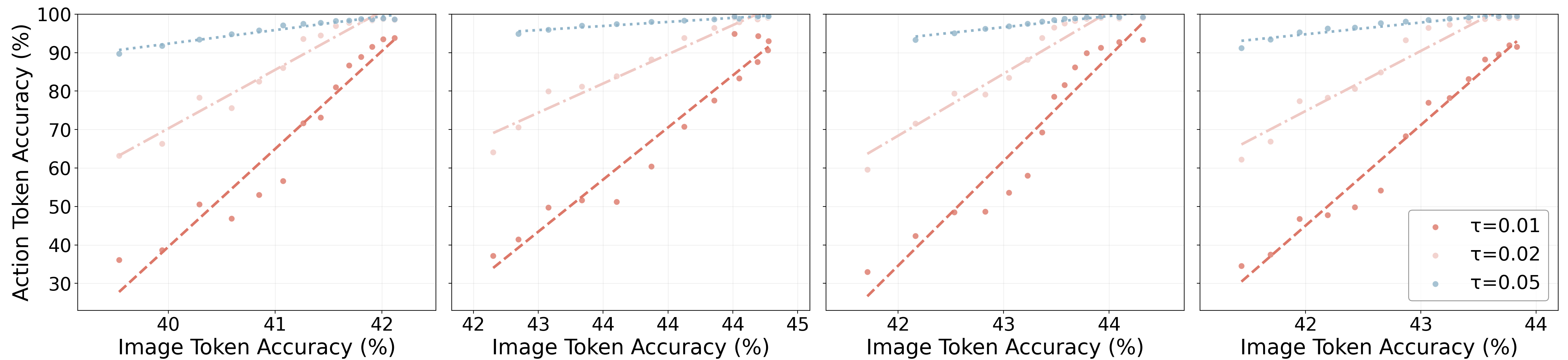}
 }
 \caption{
  \textbf{Correlation between Image and Action Token Accuracy.}
    It shows the relationship between image token accuracy and action token accuracy across the four LIBERO suites. 
    Each subfigure corresponds to one suite, and action accuracy is reported at different tolerance levels ($\tau$=0.01, 0.02, 0.05). 
    The consistent positive correlation across all settings suggests that higher-quality visual foresight is closely aligned with improved action prediction reliability.
 }
 \label{fig:img2act_corr}
\end{figure}

\cref{fig:img2act_corr} illustrates the relationship between image token accuracy and action token accuracy across four LIBERO suites. 
Across all error tolerance levels ($\tau$=0.01, 0.02, 0.05), we observe a consistent positive correlation, confirming that improvements in visual foresight are closely aligned with improvements in action prediction.
Notably, the absolute value of image token accuracy remains relatively low (around 40–45\% on average). 
This limitation arises from the fact that our model is not pretrained on large-scale generative datasets, which makes fine-grained token prediction particularly challenging. 
Nevertheless, even with imperfect image token accuracy, the generated foresight provides sufficient task-relevant cues for the action expert, leading to high action token accuracy.

These results highlight two important insights:
\begin{enumerate}[noitemsep,leftmargin=*]
    \item Pixel-level image prediction is not strictly necessary for effective action planning: even when the average image token accuracy remains modest, the generated foresight still conveys sufficient task-level cues to support high action accuracy.
    \item The strong positive correlation indicates that improvements in image token prediction are reflected in action reliability. Thus, while pixel fidelity is not required, advancing the quality of visual token prediction remains a promising way to enhance downstream action performance.
\end{enumerate}

%% file: sections/relatedwork.tex
\section{Related work}
\label{sec:related}
\subsection{Vision Langauge Action Model}
% With  in visual understanding and instruction following, researchers have sought to transfer these capabilities into embodied robots, enabling robots to complete tasks specified by natural language instructions. 
The rapid progress of multimodal large language models (MLLMs)~\citep{liu2023visualinstructiontuning,openai2024gpt4technicalreport,yang2025qwen3,bai2025qwen25vl} has motivated the development of Vision Language Action (VLA) models. 
VLA models incorporate a vision language model and augment it with an action prediction module~\citep{black2024pi_0,kim2024openvlaopensourcevisionlanguageactionmodel,qu2025spatialvlaexploringspatialrepresentations,song2025humeintroducingsystem2thinking,team2025gemini,cheang2025gr3technicalreport,bjorck2025gr00t_n1,yang2025instructvlavisionlanguageactioninstructiontuning,bu2025univlalearningacttaskcentric,qu2025embodiedonevisioninterleavedvisiontextactionpretraining}. 
They leverages the strong perceptual and linguistic grounding of pretrained VLMs, allowing robots to interpret human instructions more flexibly than purely reactive policies.
Despite this promise, current VLA models remain limited in robustness. 
Most formulations still predict actions reactively from the current state without reasoning about how the scene may evolve, leading to short-sighted behavior in dynamic and long-horizon tasks. 
Despite some studies having attempted to incorporate temporal memory~\citep{li2025cronusvlatransferringlatentmotion,shi2025memoryvlaperceptualcognitivememoryvisionlanguageaction} and post-train VLA with reinforcement learning methods~\citep{zhang2025grapegeneralizingrobotpolicy,lu2025vlarlmasterfulgeneralrobotic}, they still struggle to cope with complex scenarios.
In contrast, we propose a unified VLA model which integrates the visual foresight generation into the decision-making pipeline, thereby enhancing its robustness capability.

\subsection{Inverse Dynamics Model}
Since directly mapping the visual observation and textual instruction to action space is challenge, prior studies~\citep{deng2025graspvlagraspingfoundationmodel, hu2024vpp, liao2025genie_envisioner,zhong2025flowvlathinkingmotionvisual,cen2025worldvlaautoregressiveactionworld,wang2025unifiedvisionlanguageactionmodel,zhao2025cotvlavisualchainofthoughtreasoning, gao2025flipflowcentricgenerativeplanning} explore to enhance action prediction by injecting auxiliary intermediate representations during training, e.g., grasping poses, segmentation masks, optical flow, or future images, to guide the model toward more structured outputs. 
However, these representations are often domain-specific and do not fully leverage the potential competence of the pretrained large language model, leaving the policies brittle when deployed beyond their training distributions.
\textbf{Inverse dynamics model}~\citep{du2023learninguniversalpoliciestextguided} can extract the underlying actions from two consecutive images, thereby decrease the difficulty of mapping from image space to action space.
Recent works~\citep{black2023zeroshotroboticmanipulationpretrained,li2025unifiedvideoactionmodel, zhu2025unifiedworldmodelscoupling,cen2025worldvlaautoregressiveactionworld,zhao2025cotvlavisualchainofthoughtreasoning, wang2025unifiedvisionlanguageactionmodel, zhang2025dreamvlavisionlanguageactionmodeldreamed} attempt to decompose the decision-making task to first generate future images or videos, and then predict actions.
Nevertheless, they mostly use the future prediction objective as a regularizer when training, but seldom generate the visual guidance during the inference stage.
In contrast, our model first predicts the next frame and then predicts the action conditioned on the predicted visual foresight, improving the robustness and generalization.
% Different from the existing inverse dynamics model applied in robot manipulation,

\subsection{Unified Vision Language Model}
Building on MLLMs, recent research explores unified models that combine visual understanding and generation within a single framework. 
Early approaches~\citep{lu2023chameleonplugandplaycompositionalreasoning, zhou2024transfusionpredicttokendiffuse,xie2024showosingletransformerunify,wang2024emu3} employ discrete visual tokenization to enable joint modeling, but suffer from information loss and weakened semantics. 
While a line of works~\citep{wu2024janusdecouplingvisualencoding,pan2025transfermodalitiesmetaqueries,chen2025blip3ofamilyfullyopen,lin2025uniworldv1highresolutionsemanticencoders} adopt modular assemblies of pretrained MLLMs and diffusion models, sacrificing true unification. 
More recent efforts~\citep{deng2025emergingpropertiesunifiedmultimodal,liao2025mogaoomnifoundationmodel} introduce Mixture-of-Transformers (MoT) architectures with separate experts for text and visual generation, but still inherit the latency of diffusion and reliance on external encoders. 
However, existing unified frameworks remain centered on visual understanding and generation, leaving action outside the scope of cognitive intelligence. 
From the perspective of embodied AI, we argue that \textbf{intelligence requires not only perceiving and imagining but also interacting with the physical world.}
Compared to understanding or generation, action is inherently more complex and demanding. 
This motivates us to extend the unified paradigm toward an understanding generation action framework, enabling agents to achieve a more complete form of cognitive intelligence.

%% file: sections/conclusion.tex
\section{Conclusion and Future Work}
This paper has introduced \modelname, a pretrained Vision-Language-Action (VLA) framework that integrates goal-conditioned visual foresight into the perception–action loop. 
Building on the principle of predictive inverse dynamics, \modelname reformulates control as foresight-guided inverse dynamics, allowing actions to be derived not only from the current state but also from an anticipated visual outcome. 
Architecturally, the model adopts a Mixture-of-Transformer design with three dedicated experts for understanding, foresight generation, and action execution, while a next-scale prediction mechanism and progressive attention scheme regulate the flow of information across modules. 
To further enhance robustness and transferability, we introduced a three-stage training recipe that progressively aligns, pretrains, and adapts the experts on large-scale and task-specific robot datasets. 
Extensive experiments across simulation benchmarks and physical platforms demonstrate that \modelname consistently surpasses reactive baselines, achieving higher success rates and improved generalization in dynamic and long-horizon tasks.

Beyond the reported performance gains, this work integrates predictive  foresight with multimodal grounding in a unified VLA framework. 
The modular architecture adapts large-scale vision–language backbones to robotic control and incorporates dedicated experts for understanding, foresight generation, and action execution. 
In parallel, the progressive training scheme offers a systematic way to align and integrate these components, ensuring that foresight signals remain consistent with semantic grounding while supporting transfer across tasks and embodiments. 
This combination contributes to policies that are less dependent on purely reactive mappings and better suited to dynamic and long-horizon scenarios. 
More broadly, the study provides evidence that coupling foresight with multimodal grounding is a viable direction for advancing robust visuomotor control.

Several avenues remain for future investigation. 
Scaling foresight-driven policies to more diverse embodiments and task families, e.g., locomotion, dexterous manipulation, or multi-agent collaboration, would provide a stronger test of generality. 
Another direction is to enrich the foresight generation module with structured world models or physics-informed priors, enabling more accurate long-horizon reasoning and robustness under distributional shift. 
Integrating reinforcement learning or online adaptation strategies with foresight-guided architectures may further allow agents to refine policies beyond imitation, supporting continual improvement in open-ended environments. 
Finally, exploring how human feedback or interactive correction can be incorporated into foresight-driven policies presents an opportunity to align embodied agents more closely with human intentions.

%% file: sections/appendix.tex
\newpage

\section{Dataset Details}
\label{appendix:data_details}
Our training corpus combines large internet-scale robot datasets with curated in-house demonstrations, spanning multiple embodiments (Genie-G1, Franka, WidowX, Google Robot, ARX LIFT II), camera viewpoints (third-person and wrist/head), and frame rates (3–30 FPS). 
In total, the corpus comprises 330.9K trajectories and 73.8M frames (\cref{tab:data_statistics}).

Training proceeds in three stages. 
Pretrain Stages I–II primarily leverage the internet datasets, i.e., Agibot-World~\citep{agibotworldcontributors2025agibotworldcolosseolargescale}, OXE-Fractal~\citep{embodimentcollaboration2025openxembodimentroboticlearning}, OXE-Bridge-v2~\citep{embodimentcollaboration2025openxembodimentroboticlearning}, and LIBERO~\citep{liu2023liberobenchmarkingknowledgetransfer}, to provide broad coverage of manipulation behaviors and visual dynamics. 
Post-train Stage III adapts the model to specific skills using a smaller but higher-quality set of in-house demonstrations collected across diverse tasks, e.g., handover, sweeping, sorting, kitchen activities, and long-horizon manipulation, with Genie-G1, Franka, and ARX LIFT II. 
\begin{table}[h]
    \centering
    \resizebox{\linewidth}{!}{
        \begin{tabular}{llllllll}
            \toprule
                \textbf{Dataset} & \textbf{Source} &  \textbf{Stage} & \textbf{Embodiment} & \textbf{Camera Views}     & \textbf{FPS}  & \textbf{\# Trajs} & \textbf{\# Frames}\\
            \midrule
                Agibot-World     & Internet        &  I + II        & Genie-G1           & head $+$ left/right wrist & 30            & 187K              & 66.4M             \\
                LIBERO           & Internet        &  I + II + III  & Franka             & 3rd + wrist               & 20            & 1.7K              & 0.3M              \\
                OXE-Bridge-v2    & Internet        &  I + II + III  & WidowX             & 3rd                       & 5             & 53.2K             & 1.9M              \\
                OXE-Fractal      & Internet        &  I + II        & Google Robot       & 3rd                       & 3             & 87.2K             & 3.8M              \\
            \midrule
                Pen              & In-house    &  III           & Genie-G1           & head $+$ left/right wrist & 30            & 152               & 78.1K             \\
                Flower           & In-house    &  III           & Genie-G1           & head $+$ left/right wrist & 30            & 199               & 132.8K            \\
                Chip             & In-house    &  III           & Genie-G1           & head $+$ left/right wrist & 30            & 100               & 54.7K             \\
                Tea (Table)      & In-house    &  III           & Genie-G1           & head $+$ left/right wrist & 30            & 197               & 103.3K            \\
                Tea (Shelf)      & In-house    &  III           & Genie-G1           & head $+$ left/right wrist & 30            & 202               & 112.0K            \\
                Bread            & In-house    &  III           & Genie-G1           & head $+$ left/right wrist & 30            & 214               & 117.8K            \\
                Handover         & In-house    &  III           & Genie-G1           & head $+$ left/right wrist & 30            & 171               & 124.4K            \\
                Handover (R2H)   & In-house    &  III           & Genie-G1           & head $+$ left/right wrist & 30            & 210              & 144.5K            \\
                Sweep            & In-house    &  III           & Franka             & 3rd + wrist               & 15            & 59                & 43.8K             \\
                Sort             & In-house    &  III           & Franka             & 3rd + wrist               & 15            & 99                & 68.2K             \\
                Dynamic-Kitchen  & In-house    &  III           & ARX LIFT II        & head $+$ left/right wrist & 30            & 48                & 57.6K             \\
                Long-horizon     & In-house    &  III           & ARX LIFT II        & head $+$ left/right wrist & 30            & 129               & 347.3K            \\
            \midrule
                Total            & -               &                & -                  & -                         & -             & 330.9K            & 73.8M             \\
            \bottomrule
        \end{tabular}
    }
    \caption{
      \textbf{Data Statistics.}
    Overview of internet-scale and in-house datasets used across different training stages. Internet datasets (top) provide large-scale pretraining data across varied robots and viewpoints, while curated in-house datasets (bottom) offer high-quality demonstrations for fine-tuning. 
    In total, the combined corpus contains 330.9K trajectories and 73.8M frames.
    }
    \label{tab:data_statistics}
\end{table}

\section{Training Details}\label{appendix:training_details}
Our model training is a three-stage process, with specific hyperparameters detailed in~\cref{tab:training_args}.
Stage I focuses on learning general visual representations. It uses a large batch size of 1280 and a high learning rate of $3.0 \times 10^{-4}$ over 512K training steps. 
In Stage II, we refine the model with a larger batch size of 2880 and a constant learning rate of $5.0 \times 10^{-5}$ for 100K steps. 
This stage introduces action prediction, using a loss weight of 0.1:1 to balance generative and action losses.
For Stage III, the model is finetuned on specific downstream tasks. 
All tasks in this stage share a smaller batch size of 128 and common settings for learning rate, and loss weight. 
However, the number of training steps (or epochs) and the Action Chunk Size are specifically adjusted for each task to account for differences in data volume and task difficulty, ensuring optimal performance across the board.
\begin{table}[h]
    \centering
    \resizebox{\linewidth}{!}{
        \begin{tabular}{lllllllll}
            \toprule
                \multirow{2}[1]{*}{\textbf{Hyperparameters}} & \multirow{2}[1]{*}{\textbf{Stage I}} & \multirow{2}[1]{*}{\textbf{Stage II}} & \multicolumn{6}{c}{\textbf{Stage III}} \\
                \cmidrule{4-9}
                                                             &                             &                              & \textbf{LIBERO} & \textbf{Simpler} & \textbf{Genie-1} & \textbf{Franka} & \textbf{Dynamic} & \textbf{Long-horizon}  \\
            \midrule
                Batch Size                                   & 1280                        & 2880                         & 128    & 128    & 128    & 128    & 128    & 128 \\
                Learning Rate                                & 3.0 $\times 10^{-4}$        & 5.0 $\times 10^{-5}$         & 5.0 $\times 10^{-5}$ & 5.0 $\times 10^{-5}$ & 5.0 $\times 10^{-5}$ & 5.0 $\times 10^{-5}$ & 5.0 $\times 10^{-5}$ & 5.0 $\times 10^{-5}$ \\
                LR Scheduler                                 & Cosine                      & Constant                     & Cosine & Cosine & Cosine & Cosine & Cosine & Cosine \\
                Loss Weight (Gen:Act)                        & -                           & 0.1:1                        & 0.1:1  & 0.1:1  & 0.1:1  & 0.1:1  & 0.1:1  & 0.1:1  \\
                Training Epochs                              & -                           & -                            & -      & 10     & 40     & 40     & 40     & 60     \\
                Training Steps                               & 512K                        & 100K                         & 100K   & -      & -      & -      & -      & -      \\
            \midrule
                Und Resolution                               & 224 $\times$ 224 & 224 $\times$ 224 & 224 $\times$ 224 & 224 $\times$ 224 & 224 $\times$ 224 & 224 $\times$ 224 & 224 $\times$ 224 & 224 $\times$ 224 \\
                Gen Resolution                               & 256 $\times$ 256 & 256 $\times$ 256 & 256 $\times$ 256 & 256 $\times$ 256 & 256 $\times$ 256 & 256 $\times$ 256 & 256 $\times$ 256 & 256 $\times$ 256 \\
                \# Num Predicted Scales                      & 10                          & 4                            & 4      & 4       & 4       & 4      & 4    & 4   \\
                Action Chunk Size                            & -                           & 30                           & 4      & 8       & 50      & 50     & 50   & 50  \\
                Denoise Steps                                & -                           & -                            & 10     & 10      & 10      & 10     & 10   & 10  \\
            \bottomrule
        \end{tabular}
    }
    \caption{
      \textbf{Training recipe of \modelname.} 
      Due to significant differences in the number of demonstrations and task difficulty across downstream tasks, the settings for Stage III are not uniform.
    }
    \label{tab:training_args}
\end{table}

\section{Real-world Task Details}
\label{appendix:task_details}
We evaluate our approach across multiple robotic platforms with tasks categorized by their core manipulation requirements and complexity levels.
Our task suite spans from basic pick-and-place operations to complex long-horizon sequences, testing fundamental capabilities including precision manipulation, dual-arm coordination, human-robot interaction, and dynamic adaptation.

\subsection{Basic Pick-and-Place Manipulation}
This category covers fundamental grasping and placement tasks involving everyday household objects.
Representative instructions include: ``Put the pen from the table into the pen holder'', ``Pick up a bag of chips and place it into the basket'', and ''Pick up a bottle of black tea and place it into the shopping cart``.
The main challenges arise from the diverse physical properties of the objects, ranging from rigid items such as pens, to smooth packages like bags of chips, and deformable object such as plastic bottles, while also requiring consistent placement accuracy across target receptacles with varying constraints.
\begin{figure}[h]
 \centering
 \adjustbox{width=\linewidth,center}{
  \includegraphics{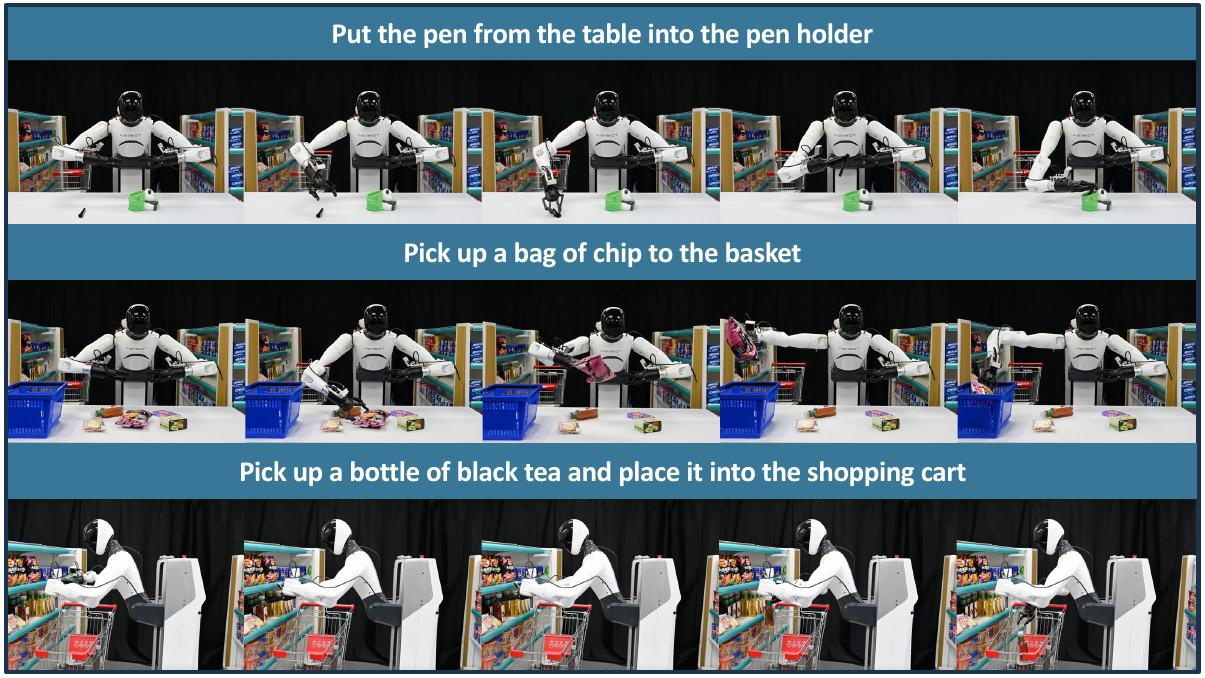}
 }
 \caption{
  \textbf{Basic Pick-and-Place Manipulation.} Examples of fundamental grasping and placement tasks across different object types and target containers. 
 }
 \label{fig:task1}
\end{figure}

\subsection{Fine-Grained Precision Manipulation}
This category evaluates the limits of robotic fine motor control through tasks demanding high precision and delicate handling.
A representative example, ``Pick up the flower and insert it into the vase'', specifically assesses fine-grained grasping and placement under tightly constrained conditions.
The challenge arises from the flower’s thin stem, which requires precise gripping to prevent damage, combined with the vase’s narrow opening, which demands accurate placement.

\begin{figure}[h]
 \centering
 \adjustbox{width=1\linewidth,center}{
  \includegraphics{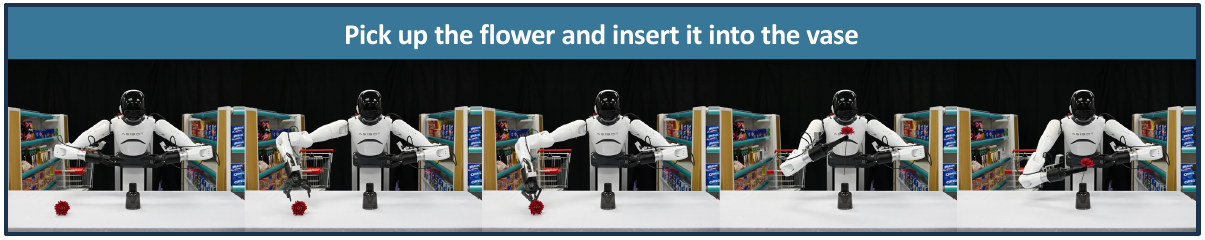}
 }
 \caption{
  \textbf{Fine-Grained Precision Manipulation.} 
  The robot is instructed to pick up the flower and insert it into the vase, which requites the precise control for handling delicate objects with narrow target constraints.
 }
 \label{fig:task2}
\end{figure}

\subsection{Dual-Arm Coordination and Human-Robot Interaction}
Bimanual manipulation tasks evaluate coordinated control of both robotic arms, focusing on spatial-temporal synchronization and inter-arm object transfer capabilities.
The primary task in this category involves the instruction ``Pick a bag of bread with the left arm, then handover, finally put it into the
basket'', and ``Pick a bottle of black tea and hand over to the person'', which requires seamless coordination between arms throughout the manipulation sequence.
\begin{figure}[h]
 \centering
 \adjustbox{width=1\linewidth,center}{
  \includegraphics{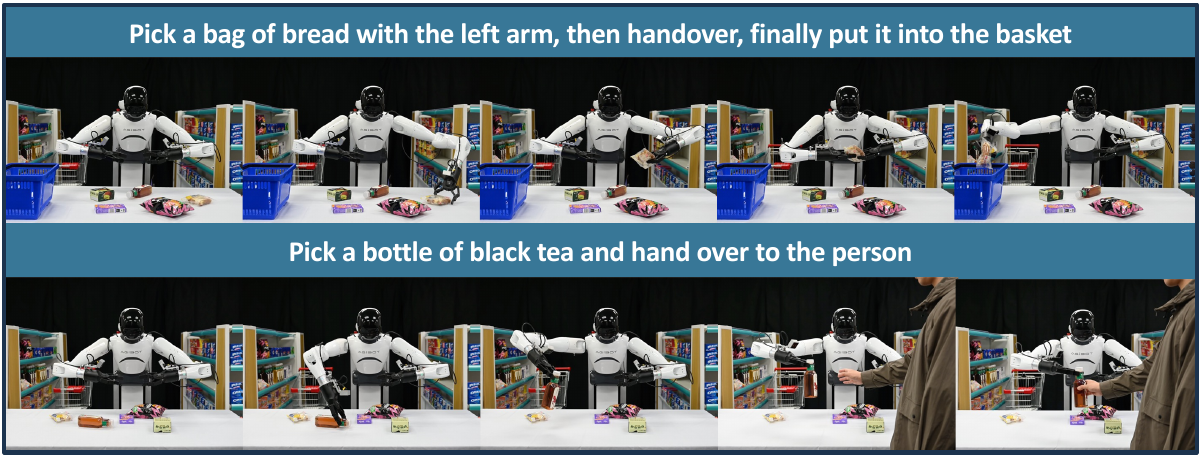}
 }
 \caption{
  \textbf{Dual-Arm Coordination and Handover.} Examples of bimanual coordination and human-robot interaction tasks demonstrating inter-arm object transfer and collaborative handover capabilities.
 }
 \label{fig:task3}
\end{figure}

\subsection{Dynamic Environment Adaptation}
Tasks in this category are performed within continuously changing environments, testing real-time tracking capabilities, and adaptive control under uncertainty. 
The instruction ``Pickup the lettuce with the right hand and then the bread with the left hand'' exemplifies the need for sophisticated trajectory prediction and real-time interception of moving targets. 
Additional challenges arise when robots must respond to unexpected dynamic events during task execution, such as objects falling or environmental conditions changing mid-operation. 

\begin{figure}[tp]
 \centering
 \adjustbox{width=1\linewidth,center}{
  \includegraphics{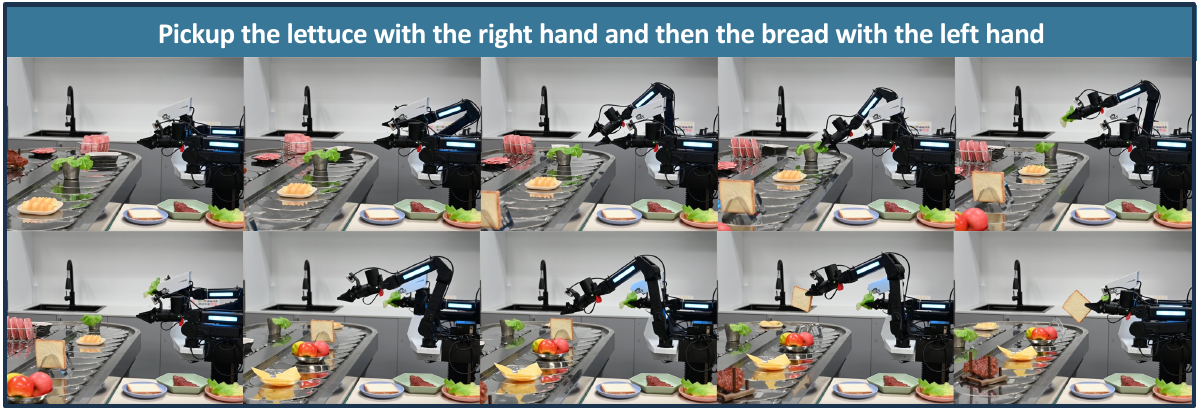}
 }
 \caption{
  \textbf{Dynamic Environment Adaptation.} Example of real-time tracking and motion prediction capabilities in continuously changing environments. The robot successfully acquires specific food items from a moving belt.
 }
 \label{fig:task4}
\end{figure}

\subsection{Long-Horizon Sequential Manipulation}
\begin{figure}[tp]
 \centering
 \adjustbox{width=\linewidth,center}{
  \includegraphics{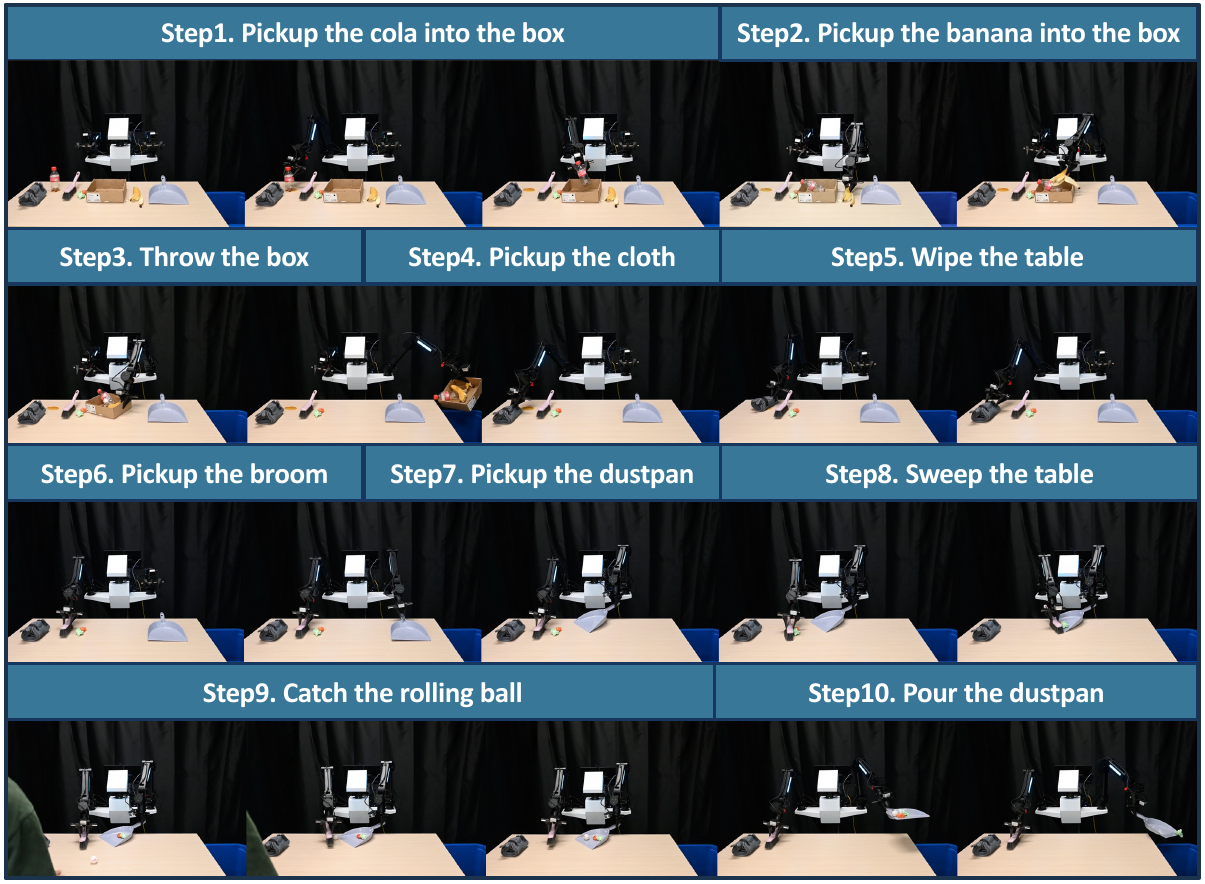}
 }
 \caption{
  \textbf{Long-Horizon Sequential Manipulation.} Example of extended multi-step task execution requiring comprehensive planning, tool usage, and task coherence maintenance. 
 }
 \label{fig:task5}
\end{figure}

Extended task sequences in this category demand comprehensive multi-step planning, coordinated tool use, and sustained task coherence across complex operation chains.
As illustrated in~\cref{fig:task5}, the key challenges include long-horizon planning over 10-step sequences with effective memory management, sequential coordination of multiple tools, handling objects with diverse physical properties—ranging from rigid and deformable to liquid materials—and dynamic replanning when unexpected events arise during execution.

\section{Deploy Platform and Latency Analysis}
All experiments are conducted on a workstation equipped with an Intel i9 CPU and an NVIDIA RTX 4090 GPU. 
All robots are connected to the host machine via wired Ethernet, thereby avoiding the additional transmission delays typically incurred in wireless communication. 
This setup ensures that the measured latency primarily reflects the computational overhead of the model itself. 
To provide a detailed view of system efficiency, we report the latency of each processing stage when the model takes three synchronized camera views as input. 
As shown in \cref{tab:deploy_platform}, the foresight generation and action decoding modules contribute the majority of the runtime, while image preprocessing and encoding remain relatively lightweight. 
Overall, the total inference time is approximately 235ms, which is sufficient for real-time deployment in embodied robotic scenarios.
\begin{table}[H]
\centering
    \begin{tabular}{c|c}
    \toprule
        \textbf{Model Part} & \textbf{Inference Time} \\
        \midrule
         image process (e.g., resize) & 18ms \\
         temporal downsampling & 28ms \\
         image encoder & 18ms \\
         foresight generation & 76ms \\
         x10 action forward pass (flow) & 95ms\\
         total inference & 235ms \\
    \bottomrule
    \end{tabular}
    \caption{
    \textbf{Latency of \modelname on the deployment platform (Intel i9 CPU + RTX 4090 GPU)}. 
    Robots are connected via wired Ethernet, so the reported numbers exclude wireless transmission delays.}
    \label{tab:deploy_platform}
\end{table}

\section{Prompt Template}\label{appendix:prompt_template}
\begin{figure}[h]
 \centering
 \adjustbox{width=1.05\linewidth,center}{
  \includegraphics{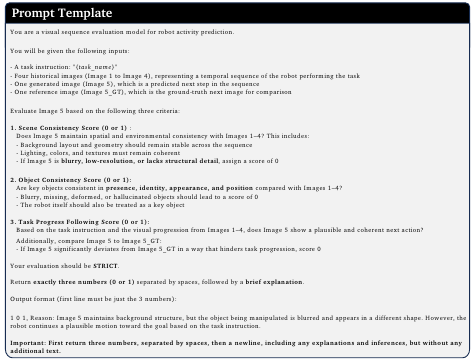}
 }
 \caption{
  \textbf{Prompt Template for Future Observation Quality Evaluation.} 
  We adopt a MLLM as evaluator, i.e., Qwen2.5-VL-32B-Instruct, and design a prompt template from three aspects to evaluate the quality of generated future observation: 1) Scene Consistency, 2) Object Consistency, and 3) Task Progress Following.
 }
 \label{fig:prompt_template}
\end{figure}

\cref{fig:prompt_template} shows the full template used for evaluating foresight image quality with Qwen2.5-VL-32B-Instruct~\citep{bai2025qwen25vl}. 
The prompt explicitly specifies the input components (task instruction, four historical frames, the predicted next-step frame, and the ground-truth frame) and guides the evaluator to assign binary scores on three aspects: (i) scene consistency, (ii) object consistency, and (iii) task progress following. 
The model is instructed to output three numbers (0/1) together with a brief explanation.
